\pdfoutput=1
\documentclass[10pt,journal,compsoc]{IEEEtran}
\usepackage{amsmath,amsfonts}
\usepackage{algorithmic}
\usepackage{algorithm}
\usepackage{array}
\usepackage[caption=false,font=normalsize,labelfont=sf,textfont=sf]{subfig}
\usepackage{textcomp}
\usepackage{stfloats}
\usepackage{url}
\usepackage{verbatim}
\usepackage{graphicx}
\usepackage{cite}
\usepackage{xspace}
\usepackage{hyperref}
\usepackage{supertabular}
\usepackage{booktabs} 
\usepackage{multirow}
\usepackage{lineno}
\usepackage{color}
\usepackage{tabularx}
\usepackage{multirow}
\newcolumntype{M}[1]{>{\raggedleft\arraybackslash}m{#1}}
\newcolumntype{T}[1]{>{\centering\arraybackslash}m{#1}}

\usepackage{bm}
\usepackage{longtable}
\usepackage{verbatim}
\usepackage{amssymb}
\usepackage{url}
\usepackage{enumitem} 
\newcommand{\paratitle}[1]{\vspace{1.2ex}\noindent\textbf{#1}}
\newcommand{\ie}{{i.e., }}

\newcommand{\model}{m-WCN\xspace}
\newcommand{\equ}{Eq.\xspace}
\newcommand{\fig}{Fig.\xspace}
\newcommand{\tab}{Tab.\xspace}

\newcommand{\mwd}{MWD\xspace}
\newcommand{\cls}{TFBC\xspace}
\newcommand{\fcl}{FCL\xspace}
\newcommand{\ftb}{FTB\xspace}
\newcommand{\highlight}{\color{black}}

\begin{document}

\title{Neuralized Multi-wavelet Decomposition for Time Series Classification and Forecasting}

\author{Xiaohan Jiang, Jingyuan Wang, Jiahao Ji, Yongyao Wang, Chen Yang, and Junjie Wu
\IEEEcompsocitemizethanks{
\IEEEcompsocthanksitem X. Jiang is with the School of Computer Science and Engineering, Beihang University, Beijing, China.
\IEEEcompsocthanksitem J. Wang is with the School of Computer Science and Engineering and the School of Economics and Management, Beihang University, Beijing, China.
\IEEEcompsocthanksitem Y. Wang, J. Ji, and C. Yang are with the School of Computer Science and Engineering, Beihang University, Beijing, China, and are also with the MOE Engineering Research Center of Advanced Computer Application Technology, Beihang University, China.
\IEEEcompsocthanksitem J. Wu is with the School of Economics and Management, Beihang University, Beijing 100191, China, and the Key Laboratory of Data and Decision Intelligence (Beihang University), Ministry of Industry and Information Technology, Beijing 100191, China.
\IEEEcompsocthanksitem Corresponding author: J. Wang (jywang@buaa.edu.cn).
}
}


\IEEEtitleabstractindextext{%
\begin{abstract}
Time series analysis is fundamental in domains such as finance, healthcare, and meteorology. Real-world time series often exhibit multiscale characteristics shaped by diverse latent factors, resulting in intricate temporal patterns and rich frequency structures. However, existing approaches typically focus on either frequency-domain decomposition or time-domain pattern extraction in isolation, neglecting their joint structure. This decoupled modeling limits representation expressiveness and undermines performance in tasks requiring simultaneous temporal and spectral reasoning. To address this gap, we propose \model, a novel end-to-end deep learning framework that neuralizes multi-wavelet decomposition for joint extraction of temporal patterns and frequency components. By approximating the classical GHM multi-wavelet transform with trainable convolutional operators and enforcing orthogonality constraints, \model produces interpretable multi-resolution representations. Built on this foundation, we introduce two task-specific architectures: \cls for time series classification, which boosts discriminative features across frequency scales, and \ftb for forecasting, which ensembles {frequency-aware predictors.}
Extensive experiments on 64 UCR datasets and {seven} public forecasting benchmarks demonstrate the effectiveness of our approach. Built on the neuralized \model, our \cls and \ftb outperform various baseline models across diverse datasets, achieving average improvements of 19.97\% in classification and 19.92\% in forecasting tasks.
\end{abstract}

}

\maketitle
\IEEEpeerreviewmaketitle

\section{Introduction}

\IEEEPARstart{A} time series is a sequence of data points recorded in temporal order. Time series analysis, including time series classification (TSC) and time series forecasting (TSF)~\cite{wang2018multilevel}, plays a crucial role in understanding dynamic systems and supporting decision-making across diverse domains such as finance, healthcare, and meteorology~\cite{wang2019alphastock, ren2022generative, duchon2012time}.

In real-world scenarios, time series are often shaped by multiple latent factors, giving rise to signals with intricate temporal patterns and diverse frequency components. For instance, from a frequency-domain perspective, traffic flow data may contain low-frequency trends reflecting daily commuting patterns and high-frequency fluctuations caused by sudden events. From a temporal perspective, the same data can be segmented into distinct intervals such as morning rush hour, evening rush hour, and off-peak periods. Effectively modeling and disentangling both temporal and frequency characteristics is therefore essential for achieving robust, interpretable, and high-quality time series analysis. To this end, decomposition-based approaches have proven effective for disentangling these complex structures~\cite{godfrey2017neural, zhang2017application}. By decomposing a time series into distinct subcomponents, such methods support more expressive representations, facilitate task-specific learning, and improve overall performance on both classification and forecasting problems.

In the literature, decomposition-inspired time series analysis methods are broadly categorized into two groups: \textit{frequency-based} and \textit{time-based} approaches. Frequency-based methods employ spectral decomposition techniques to transform a time series into a frequency domain representation, where each component captures information associated with a specific frequency. Common techniques in this category include the Discrete Fourier Transform (DFT)~\cite{dft}, Discrete Wavelet Transform (DWT)~\cite{mallat1989theory}, and Z-transform~\cite{palani2022z}. These methods are grounded in strong mathematical foundations and offer interpretable decompositions that reveal intrinsic properties of the input signals.
Despite their theoretical rigor, frequency-based methods often rely on a fixed and limited set of basis functions, which may impose overly restrictive assumptions about the underlying structure of the time series. As a result, they may struggle to effectively capture complex patterns, local variations, and nonstationary behaviors commonly found in real-world data.

The second category comprises time-based approaches, which treat a time series as an ordered sequence of data points and aim to uncover patterns directly through data-driven techniques. These methods typically decompose a series into multiple temporal segments or components, each representing a distinct underlying pattern. Representative techniques include Empirical Mode Decomposition (EMD)~\cite{rilling2003empirical} and shapelet-based methods~\cite{ye2009time, hills2014classification, ye2011time}. While time-based approaches offer flexibility and are well-suited to capturing localized or irregular patterns, they often neglect the frequency characteristics of the signal. This omission can limit their capacity to detect meaningful structures that manifest across different frequency scales, potentially compromising performance in tasks where frequency dynamics are critical.

The emergence of deep learning has introduced powerful tools for advancing decomposition-based time series analysis. Early approaches typically use frequency or temporal decomposition as a preprocessing step to extract handcrafted features, which are then fed into deep neural networks~\cite{zhou2022fedformer, hajiabotorabi2019improving, liu2013forecasting}. Although this loosely coupled paradigm can enhance performance compared to using raw time series, it suffers from a lack of end-to-end optimization, as the decomposition and learning stages are treated as separate processes with independently trained parameters. More recently, efforts have been made to incorporate decomposition techniques directly into end-to-end deep learning architectures~\cite{wang2018multilevel, wang2023when}. These integrated models aim to jointly optimize decomposition and representation learning. However, most of these methods primarily focus on frequency-domain decomposition, often overlooking the rich structural information embedded in temporal patterns. As a result, they fall short of fully capturing the complementary insights offered by both time- and frequency-domain representations. Therefore, there is a critical need for a unified framework that seamlessly integrates both time-domain pattern decomposition and frequency-domain component analysis within a learnable, end-to-end neural architecture. Such a framework should leverage the interpretability and structural rigor of traditional decomposition methods while incorporating the adaptive learning capabilities of deep networks.

To address these challenges, we propose \underline{m}ulti-\underline{W}avelet \underline{C}onvolution \underline{N}etworks (\model), a novel framework that performs simultaneous temporal pattern and frequency decomposition in a fully trainable, end-to-end manner. Our approach is grounded in the theory of Multi-Wavelet Decomposition (\mwd), which we leverage to construct a pattern-frequency joint decomposition architecture. Specifically, \mwd first transforms the input time series into multiple temporal pattern components via a pre-filtering step, and then further decomposes each pattern component into hierarchical frequency components using dedicated scaling and wavelet functions.
To preserve the mathematical rigor of \mwd while enabling data-driven flexibility, we neuralize the two core steps of the GHM multi-wavelet algorithm — an established instance of \mwd — using convolutional neural networks. In the temporal pattern decomposition phase, \model enhances the fixed GHM pre-filter with learnable convolutional kernels to adaptively extract representative time-domain patterns. In the frequency decomposition phase, \model enhances the GHM scaling and wavelet functions using trainable two-dimensional convolutional kernels, enabling it to better capture frequency structures tailored to the input data. Unlike traditional multi-wavelet decomposition methods with fixed parameters, all components in \model are fully learnable and can be fine-tuned to fit the training data for various learning tasks. This enables \model to combine the theoretical advantages of multi-wavelet signal decomposition with the powerful representation learning capabilities of deep neural networks. Furthermore, we introduce an orthogonal regularization term to promote diversity among learned components, thereby preserving the orthogonality property intrinsic to \mwd and ensuring interpretability of the extracted patterns.

Based on \model, we also propose two task-specific deep learning models for the time series classification (TSC) and time series forecasting (TSF), respectively. The key issue in TSC is to extract discriminative features from a time series. Therefore, we propose a \model-based Time-Frequency Boosting Classification (\cls) model, which leverages frequency components extracted by \model in a coarse-to-fine boosting manner, where higher-frequency features are used to complement the information missed by lower-frequency ones. This design enables the model to exploit complementary time-frequency cues, enhancing its discriminative power for time series classification. For the TSF task, a key challenge lies in accurately modeling future dynamics by capturing latent trends across multiple frequency scales. { To tackle this, we propose the Frequency {TSMixer}~\cite{chen2023tsmixer} Bagging (\ftb) model, which processes each frequency component extracted by \model using a dedicated {TSMixer} network.} The outputs from all frequency-specific {TSMixer} are then aggregated in a bagging manner to generate the final forecast, effectively leveraging complementary information across different frequency bands. To facilitate more effective model training, we design tailored pre-training strategies for both \cls and \ftb. For \cls, we introduce a Frequency Contrastive Learning (FCL) objective that encourages consistency among representations across frequency components of the same input. For \ftb, we propose a Frequency Representation Pre-training (FRP) strategy, which guides each {predictor} to predict its corresponding future frequency component, thereby enhancing the frequency-awareness and forecasting capability of the model.

We evaluate the effectiveness of \model and its task-specific variants through extensive experiments across diverse benchmarks. We evaluate \cls on 64 UCR time series datasets for TSC, and \ftb on {seven} real-world public datasets for TSF.
The results demonstrate the two models' superiority to various baselines by an average performance improvement of 19.97\% and 19.92\% in classification and forecasting tasks, respectively (See Sec.~4 of the Supplementary Materials for the calculation details of average performance improvement).
The contributions of our work can be summarized as follows:
\begin{itemize}[leftmargin=*]
    \item We propose \model, the first end-to-end deep learning framework that jointly integrates frequency and temporal-pattern decomposition via a neuralized multi-wavelet design, bridging classic signal processing and modern representation learning.
    \item We develop two task-specific architectures: \cls for classification and \ftb for forecasting. \cls employs frequency-wise boosting to capture complementary decision cues across scales, while \ftb performs frequency-aligned forecasting by assigning dedicated predictors to each frequency band. Both architectures demonstrate strong adaptability to the structural characteristics of time series in their respective tasks.
  \item The proposed models achieved the state-of-the-art performance over a large number of real-world datasets.
\end{itemize}




\section{Related Work}\label{sec:relate}

{{\bf Pattern Analysis (PA).} It is a crucial technique for identifying distinctive data properties \cite{rilling2003empirical,ye2009time}. In time series analysis, PA helps extract informative features to support downstream models \cite{hills2014classification,wang2020deep}. Several deep learning methods have successfully integrated PA with neural networks, achieving notable results. Examples include empirical mode decomposition \cite{emd2017}, tensor decomposition \cite{wang2019understanding}, and shapelet-based methods \cite{grabocka2014learning}. However, many of these methods overlook frequency information, which can limit their ability to capture meaningful patterns.}

{{\bf Frequency Analysis.} It is a key technique for revealing data characteristics in the frequency domain, using methods such as Discrete Fourier Transform (DFT) \cite{dft} and Discrete Wavelet Transform (DWT) \cite{mallat1989theory}. In time series analysis, traditional methods typically incorporate frequency coefficients from discrete analyses as model features but lack deeper integration and refinement \cite{middlehurst2021hive}. Deep learning has broadened the use of frequency analysis—neural networks can automatically tune key frequency coefficients \cite{wang2018multilevel}, and neural operator learning has been combined with frequency analysis \cite{gupta2021multiwavelet}. Self-attention mechanisms have also been used to integrate DFT and DWT \cite{zhou2022fedformer}.
However, existing approaches often fail to effectively combine frequency-based and pattern-based methods, both of which are critical in time series analysis.
The Discrete Multi-Wavelet Transform (DMWT) \cite{xia1996design, tham2000general} extends DWT by integrating frequency and pattern information to extract joint features. However, DMWT cannot adapt pattern modes to varying data characteristics, often producing patterns that misalign with the actual data. Our approach addresses this by using deep neural networks to learn and extract combined frequency-pattern features that better reflect the data's intrinsic regularities.}

{\bf Time Series Classification (TSC).} TSC aims to categorize time series patterns using models trained on labeled data. Traditional methods, including distance-based \cite{sequencetsc,dtw}, feature extraction \cite{Bagofwords}, and ensemble approaches \cite{middlehurst2021hive}, often rely on handcrafted features like distance metrics and differences. However, these techniques may struggle with large or complex datasets.
Recently, deep neural networks have become a powerful tool for TSC, capable of automatically learning complex features. This includes supervised feature mining \cite{grabocka2014learning}, unsupervised feature learning \cite{franceschi2019unsupervised}, and Transformer-based models \cite{transformer_tst,chowdhury2022tarnet}. While these models effectively generate diverse features through representational learning, they may still overlook unique aspects of time series data, such as mixed Pattern-Frequency features.

{{\bf Time Series Forecasting (TSF).}
TSF refers to predicting future values of a time series using past and present data, which is widely adopted in nearly all application domains. A classic model is autoregressive integrated moving average (ARIMA) \cite{arima}, with a great many variants, such as ARIMA with explanatory variables (ARIMAX) \cite{kongcharoen2013autoregressive} and seasonal ARIMA (SARIMA) \cite{SARIMA}, to meet the requirements of various applications.
In recent years, deep learning has emerged as the leading approach in this field. It includes various methodologies such as neural ordinary differential equations \cite{ODE_1,ji2022stden}, probabilistic forecasting models \cite{deepar}, and transformer-based architectures \cite{wu2021autoformer,han2025bridging}. Despite these advancements, integrating both pattern and frequency features remains a challenge, with few studies successfully addressing this dual consideration.}

\section{Preliminaries}\label{sec:prelim}

\subsection{Notations}

In this paper, we use lowercase letters in regular font ($a, b$) to denote scalars, lowercase bold letters ($\bm{a}, \bm{b}$) to denote vectors, uppercase bold letters ($\bm{A}, \bm{B}$) to denote matrices, and uppercase calligraphic letters ($\bm{\mathcal{A}}, \bm{\mathcal{B}}$) to denote tensors or sets of matrices. The uppercase  letters in regular font ($A, B$) are used to denote constants.

For a matrix {\small $\bm{A} \in \mathbb{R}^{I \times J}$}, its $i$-th row vector and $j$-th column vector are denoted as $\bm{a}_{i:}$ and $\bm{a}_{:j}$, respectively. For a third-order tensor {\small $\bm{\mathcal{A}} \in \mathbb{R}^{I \times J \times K}$}, its horizontal slices, lateral slices, and frontal slices are denoted as $\bm{A}_{i::}$, $\bm{A}_{:j:}$, and $\bm{A}_{::k}$, respectively. The row, column, and tube fibers (vectors) of the tensor are denoted as $\bm{a}_{i:k}$, $\bm{a}_{:jk}$, and $\bm{a}_{ij:}$, respectively.

\begin{table}[!t]
  \centering
  \caption{Notations used in multi-wavelet decomposition and \model.}\vspace{-3mm}
  \resizebox{0.9\linewidth}{!}{
    \setlength{\tabcolsep}{0.8mm}
    \renewcommand{\arraystretch}{1}
  \begin{tabular}{l|l}
    \toprule
    {\bf Notations} & {\bf Description} \\ \midrule
    $\bm{s} = (s_0, \ldots, s_{T})$ & The input series for \model and TSC. \vspace{0.05cm}\\
    $\bm{c} = (c_1, \ldots, c_M)$ & The one-hot category label for TSC. \vspace{0.05cm}\\
    $\bm{s}_t = (s_{t-L}, \ldots, s_{t})$    & The input series for \model and TSF. \vspace{0.05cm}\\
    $\bm{z}_t = (s_{t+1}, \ldots, s_{t+L'})$ & The series to be predicted for TSF. \\ \midrule
    $\bm{X}^{h_n}, n \in \{1, \ldots, N\}$  &  The $n$-th high frequency component of $\bm{s}$.\vspace{0.05cm}\\
    $\bm{X}^{l_n},\; n \in \{1, \ldots, N\}$   &  The $n$-th low frequency component of $\bm{s}$.\vspace{0.05cm}\\
    $\bm{Z}_t^{h_n}$ and $\bm{Z}_t^{l_N}$ & The frequency components of $\bm{z}_t$ in TSF. \vspace{0.05cm}\\ \midrule
   \multirow{2}{*}{$\psi(t)$ and $\phi(t)$} & The wavelet and scaling function of    \\
   & { single wavelet decomposition.}  \vspace{0.05cm}\\
   \multirow{2}{*}{$\left\{\psi_1(t), \ldots, \psi_K(t)\right\}$} & The wavelet functions of  multi-wavelet \\
   & { decomposition.}  \vspace{0.05cm}\\
   \multirow{2}{*}{$\left\{\phi_1(t), \ldots, \phi_K(t)\right\}$} & The scaling functions of multi-wavelet \\
   & { decomposition.}  \vspace{0.05cm} \\
   \multirow{2}{*}{$\bm{\mathcal{H}}$ and $\bm{\mathcal{L}}$} & The wavelet and scaling function of the  \\
   & { GHM decomposition.} \vspace{0.05cm}\\
   \multirow{2}{*}{$\tilde{\bm{\mathcal{H}}}$ and $\tilde{\bm{\mathcal{L}}}$} & The wavelet and scaling function of the \\
   & { \model model.} \vspace{0.05cm}\\\midrule
   $\bm{g}^{l_n}$ and $\bm{g}^{h_n}$ &  The representations of frequency in TSC. \vspace{0.05cm} \\
    $\bm{E}_t^{l_n}$ and $\bm{E}_t^{h_n}$ & The representations of frequency in TSF.  \vspace{0.05cm} \\
    $\bm{\Theta}$, $\bm{\mathcal{W}}$ & Learnable parameters. \vspace{0.05cm} \\
       \bottomrule
  \end{tabular}}\vspace{-4mm}\label{eq:notations}
\end{table}

\tab~\ref{eq:notations} summarizes the notations used for multi-wavelet decomposition and our \model framework. In this paper, subscripts (e.g., $s_t$) are used to denote the index of a variable within a series, vector, matrix, or tensor, while superscripts (e.g., $\bm{X}^{l_n}$) are used to indicate the corresponding frequency components. Specifically, $*^{l_n}$ denotes the low-frequency component at the $n$-th layer, and $*^{h_n}$ denotes the high-frequency component at the $n$-th layer.

\subsection{Wavelet Decomposition}\label{sec:wavelet}

Given a series $\bm{s} = (s_0, \ldots, s_t, \ldots, s_T)$, wavelet decomposition applies a wavelet function $\psi(t)$ and a scaling function $\phi(t)$ to extract its high- and low-frequency components:
\begin{equation}\label{eq:single_wavelet}\small
    s_t^h = \sum_{i=0}^{T} \psi(t) s_i,\;\;\;\mathrm{and}\;\;\;\; s_t^l = \sum_{i=0}^{T} \phi(t) s_i.
\end{equation}
Here, {\small $\bm{s}^h = (s_0^h, \ldots, s_t^h, \ldots, s_T^h)$} and {\small $\bm{s}^l = (s_0^l, \ldots, s_t^l, \ldots, s_T^l)$} denote the high-frequency and low-frequency component series, respectively. The decomposition in Eq.~\eqref{eq:single_wavelet} effectively captures frequency information from the input series: high-frequency components reflect short-term variations, while low-frequency components capture long-term trends. Different wavelet decomposition methods employ various choices of wavelet and scaling functions~\cite{Lee2019}.

\subsection{Multi-wavelet Decomposition}\label{sec:multi-wavelet}

In the basic wavelet decomposition described in Sec.~\ref{sec:wavelet}, the input is a scalar time series, and the algorithm applies a single wavelet function $\psi(t)$ and a single scaling function $\phi(t)$ to extract frequency components. This method, known as single wavelet decomposition, assumes that the series can be characterized by a single set of frequency components. However, real-world time series often exhibit complex and rich spectral structures, sometimes requiring decomposition into many frequency components. To better capture such complexity, Multi-Wavelet Decomposition (\mwd)~\cite{keinert2003wavelets} provides a more expressive and flexible framework for frequency analysis. The \mwd process consists of two key steps: a time-domain pre-filtering step and a frequency-domain decomposition step.

\paratitle{Pre-filtering (Time-domain Pattern Decomposition).}
In the pre-filtering step, \mwd transforms the scalar time series into a multivariate (vector) time series. Given a sequence $\bm{s} = (s_1, \ldots, s_t, \ldots, s_T)$, a linear transformation is applied to construct the vector series:
\begin{equation}\label{eq:prefilterequation}\small
   \bm{x}_{t} = \bm{P} \cdot (s_t, s_{t+1}, \ldots, s_{t+M})^\top,
\end{equation}
where {\small $\bm{P} \in \mathbb{R}^{K \times (M+1)}$} is a filter matrix that maps a local segment {\small $(s_t, s_{t+1}, \ldots, s_{t+M})^\top$} into the vector {\small $\bm{x}_{t} \in \mathbb{R}^K$} at time slice $t$. The filter matrix $\bm{P}$ is predefined according to the specific multi-wavelet decomposition algorithm~\cite{xia1996design}. Over all $T$ time slices, this transformation produces a $K$-dimensional multivariate series {\small $\bm{X} = (\bm{x}_{1}, \ldots, \bm{x}_{t}, \ldots, \bm{x}_{T})$}~\footnote{We apply padding to handle the insufficient parts at the end of the sequence, ensuring that the pre-filtering operation can be performed consistently across all time steps.}. Each subseries {\small $\bm{x}_{k:} = (x_{k1}, \ldots, x_{kT})$} in {\small $\bm{X}$} represents a projection of the original series onto a distinct subspace. In this way, the pre-filtering process extracts multiple time-domain patterns {\small $\bm{x}_{k:}, k\in \{1, \ldots, K\}$ } from the original input series $\bm{s}$, capturing diverse temporal structures within the data.



\paratitle{Wavelet Decomposition (Frequency-domain Component Decomposition).}
In the wavelet decomposition step, \mwd utilizes multiple wavelet and scaling functions to process different subseries of $\bm{X}$. Given $K$ wavelet functions {\small $\{\psi_1(t), \ldots, \psi_K(t)\}$} and scaling functions {\small $\{\phi_1(t), \ldots, \phi_K(t)\}$}, the vector series $\bm{X}$ is transformed as follows:
\begin{equation}\label{eq:multi_decomposition}\scriptsize
  \begin{pmatrix}
    \tilde{x}_{1t}^h  \\
    \vdots \\
    \tilde{x}_{Kt}^h
  \end{pmatrix}
    =
  \begin{pmatrix}
    \sum_{i=0}^{T} \psi_1(t) x_{1i}\\
    \vdots \\
    \sum_{i=0}^{T} \psi_K(t) x_{Ki}
  \end{pmatrix},\quad
  \begin{pmatrix}
    \tilde{x}_{1t}^l  \\
    \vdots \\
    \tilde{x}_{Kt}^l
  \end{pmatrix}
    =
  \begin{pmatrix}
    \sum_{i=0}^{T} \phi_1(t) x_{1i}\\
    \vdots \\
    \sum_{i=0}^{T} \phi_K(t) x_{Ki}
  \end{pmatrix}.
\end{equation}
Letting {\small $\tilde{\bm{x}}_{t}^h = (\tilde{x}_{1t}^h, \ldots, \tilde{x}_{Kt}^h)^\top$} and {\small $\tilde{\bm{x}}_{t}^l = (\tilde{x}_{1t}^l, \ldots, \tilde{x}_{Kt}^l)^\top$}, we define {\small $\tilde{\bm{X}}^h = (\tilde{\bm{x}}_{1}^h, \ldots, \tilde{\bm{x}}_{T}^h)$} and {\small $\tilde{\bm{X}}^l = (\tilde{\bm{x}}_{1}^l, \ldots, \tilde{\bm{x}}_{T}^l)$} as the high- and low-frequency representations of the vector series {\small $\bm{X}$}.

To preserve the total length of the decomposed components relative to the original input sequence $\bm{X}$, a downsampling operation is applied to both frequency representations:
\begin{equation}\label{eq:down_sampling}\small
   \bm{X}^h = \tilde{\bm{X}}^h \downarrow 2,\;\;\; \bm{X}^l = \tilde{\bm{X}}^l \downarrow 2,
\end{equation}
where $\downarrow 2$ denotes a downsampling operation that reduces the sequence length by half. The resulting sequences {\small $\bm{X}^h \in \mathbb{R}^{K \times (T/2)}$} and {\small $\bm{X}^l \in \mathbb{R}^{K \times (T/2)}$} represent the high- and low-frequency components, respectively. Each subseries {\small $\bm{x}_{k:}^h$} and {\small $\bm{x}_{k:}^l$} in {\small $\bm{X}^h$} and {\small $\bm{X}^l$} corresponds to the frequency-domain transformation of a distinct time-domain pattern identified during the pre-filtering step.

The \mwd recursively applies the decomposition process defined in Eq.~\eqref{eq:multi_decomposition} and Eq.~\eqref{eq:down_sampling}. Letting {\small $\bm{X}^{l_0} = \bm{X}$} denote the input to the first layer, the high- and low-frequency components at the $n$-th layer are computed as:
\begin{equation}\small
  \left\{\bm{X}^{h_n}, \bm{X}^{l_n}\right\} = \mathrm{MWD}\left(\bm{X}^{l_{n-1}}\right),
\end{equation}
where $\mathrm{MWD}(\cdot)$ represents the composite operation of multi-wavelet transformation and downsampling.

For an $N$-layer decomposition, the final output is a set of sequences {\small $\{\bm{X}^{h_1}, \bm{X}^{h_2}, \ldots, \bm{X}^{h_N}, \bm{X}^{l_N}\}$}, where {\small $\bm{X}^{h_1}, \ldots, \bm{X}^{h_N}$} are the high-frequency components from each layer, and {\small $\bm{X}^{l_N}$} is the low-frequency component obtained at the final layer. Intermediate low-frequency components are used only for recursive decomposition and are not retained in the final output.

\begin{figure}[!t]
    \centering
    \includegraphics[width=0.8\columnwidth]{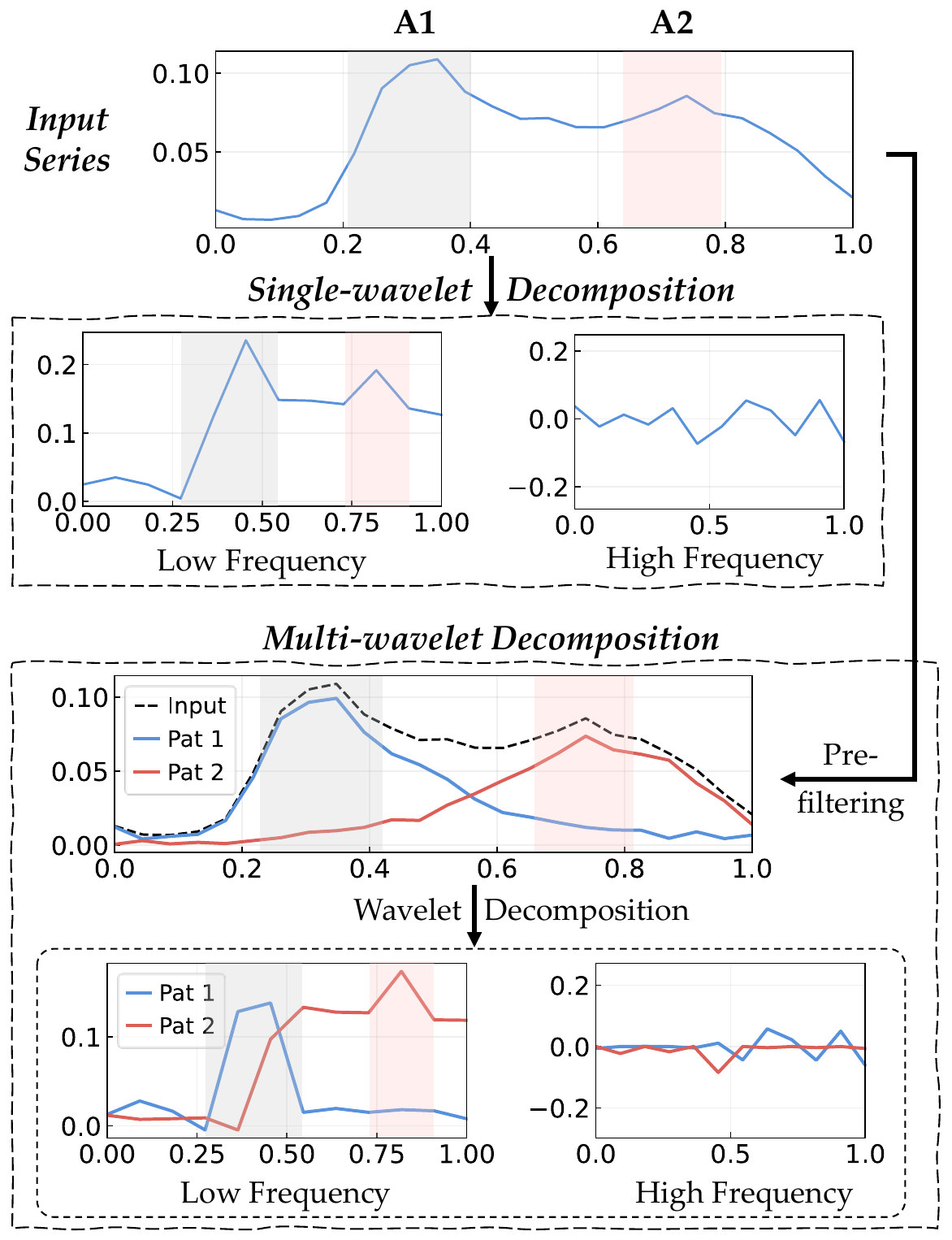}
    \vspace{-3mm}
    \caption{\highlight Comparison of single-wavelet and multi-wavelet decomposition (Pat: Pattern). { ${\bf A1}$ and ${\bf A2}$ denote the temporal patterns associated with the morning peak and the evening peak, respectively.} Single-wavelet decomposition processes the mixed patterns ${\bf A1}$ and ${\bf A2}$ together, whereas multi-wavelet decomposition first separates them through pre-filtering and then performs frequency decomposition on the separated pattern components.}
    \label{fig:motivation}\vspace{-4mm}
\end{figure}

{\vspace{1.2ex} \bf \em Remark:} Compared with single wavelet decomposition, multi-wavelet decomposition offers significant advantages in analyzing temporal compound characteristics within sequential signals. Fig.~\ref{fig:motivation} illustrates this with an example of total traffic volume in a city over a single day. The series exhibits two distinct patterns—morning and evening peaks—representing compound characteristics. In single wavelet decomposition, both peaks must be analyzed simultaneously using the same wavelet basis, which may result in mixed or entangled frequency components. In contrast, multi-wavelet decomposition first applies a pre-filtering step to separate these patterns and then analyzes their frequency components using different wavelet and scaling functions. This leads to more homogeneous subseries, making them easier to model and predict in downstream tasks.

\section{Multi-Wavelet Convolution Network}\label{sec:MWDCN}

In this section, we introduce a novel network architecture, \model (\underline{m}ulti-\underline{W}avelet \underline{C}onvolution \underline{N}etworks). The core idea of \model is to use a convolutional network structure to neuralize the pre-filtering and decomposition steps of multi-wavelet decomposition. Sec.~\ref{sec:GHM} presents the backbone multi-wavelet algorithm adopted by \model. Sec.~\ref{sec:pre_module} details the neuralization of the pre-filtering step, and Sec.~\ref{sec:decomp_module} describes the neuralization of the decomposition step.

\subsection{GHM Multi-wavelet and \model Framework}\label{sec:GHM}

The proposed \model adopts the Geronimo--Hardin--Massopust (GHM) multi-wavelet as its backbone decomposition algorithm. GHM is a classical multi-wavelet transform, known for its compact support, orthogonality, and favorable regularity properties of its scaling and wavelet functions~\cite{geronimo1994fractal}.

In the time-domain pattern decomposition step, GHM employs a predefined projection matrix to transform the original scalar series into a two-dimensional vector series. This projection matrix $\bm{P}$, used in Eq.~\eqref{eq:prefilterequation}, is defined as:
\begin{equation}\label{eq:qmatrix}\scriptsize
\bm{P} =
\begin{pmatrix}
-0.0036 & 0.0844 & 0.9928 & 0.0849 & -0.0036 & 0.00015 \\
0.00015 & -0.0036 & -0.0849 & 0.9928 & -0.0844 & 0.0036
\end{pmatrix}.
\end{equation}
By applying the projection matrix $\bm{P}$, the GHM algorithm transforms the original scalar input series into a two-dimensional vector series, preparing it for subsequent frequency-domain decomposition.

For the wavelet decomposition step, GHM defines the scaling functions using four matrices:
\begin{equation}\label{eq:lmatrix}\small
\begin{aligned}
  &\bm{L}_{:1:} =
  \begin{pmatrix}
      \frac{3}{5\sqrt{2}} & -\frac{1}{20} \\
      0 & \frac{9}{20}
  \end{pmatrix}, \;\;
  &&\bm{L}_{:2:} =
  \begin{pmatrix}
      \frac{4}{5} &-\frac{3\sqrt{2}}{10} \\
      0 & -\frac{3\sqrt{2}}{10}
  \end{pmatrix}, \\
  &\bm{L}_{:3:} =
  \begin{pmatrix}
      \frac{3}{5\sqrt{2}} & \frac{9}{20} \\
      0 & -\frac{1}{20}
  \end{pmatrix}, \;\;
  &&\bm{L}_{:4:} =
  \begin{pmatrix}
      0 & \frac{1}{\sqrt{2}} \\
      0 & 0
  \end{pmatrix}.
\end{aligned}
\end{equation}
These matrices form a third-order tensor {\small $\bm{\mathcal{L}} =$ $\big(\bm{L}_{:1:},$ $\bm{L}_{:2:},$ $\bm{L}_{:3:},$ $\bm{L}_{:4:}\big)$}, which serves as a low-pass filter. The horizontal slices {\small $\bm{L}_{1::} \in \mathbb{R}^{4 \times 2}$} and {\small $\bm{L}_{2::} \in \mathbb{R}^{4 \times 2}$} of {\small $\bm{\mathcal{L}}$} correspond to the two scaling functions of the GHM algorithm, i.e., {\small $\phi_1 = \bm{L}_{1::}$} and {\small $\phi_2 = \bm{L}_{2::}$}. 
Similarly, the wavelet functions are defined by the following matrices:
\begin{equation}\label{eq:hmatrix}\small
\begin{aligned}
  &\bm{H}_{:1:} =
  \begin{pmatrix}
      -\frac{1}{20} & \frac{1}{10\sqrt{2}} \\
      \frac{9}{20}  &\frac{9}{10\sqrt{2}}
  \end{pmatrix}, \;\;
  &&\bm{H}_{:2:} =
  \begin{pmatrix}
     -\frac{3\sqrt{2}}{10} &  \frac{3}{10}  \\
     -\frac{3\sqrt{2}}{10}  & -\frac{3}{10}
  \end{pmatrix}, \\
  &\bm{H}_{:3:} =
  \begin{pmatrix}
      \frac{9}{20} & -\frac{9}{10\sqrt{2}} \\
      -\frac{1}{20} & -\frac{1}{10\sqrt{2}}
  \end{pmatrix}, \;\;
  &&\bm{H}_{:4:} =
  \begin{pmatrix}
      -\frac{1}{\sqrt{2}}  & 0 \\
      0 & 0
  \end{pmatrix}.
\end{aligned}
\end{equation}
These matrices form the high-pass filter tensor {\small $\bm{\mathcal{H}} =$ $\big(\bm{H}_{:1:},$ $\bm{H}_{:2:},$ $\bm{H}_{:3:},$ $\bm{H}_{:4:}\big)$}. The horizontal slices {\small $\bm{H}_{1::} \in \mathbb{R}^{4 \times 2}$} and {\small $\bm{H}_{2::} \in \mathbb{R}^{4 \times 2}$} correspond to the two wavelet functions of the GHM algorithm, denoted as {\small $\psi_1 = \bm{H}_{1::}$} and {\small $\psi_2 = \bm{H}_{2::}$}. 

In the GHM algorithm, {\small $\bm{\mathcal{L}}$} and {\small $\bm{\mathcal{H}}$} are used as convolution kernels to extract the low- and high-frequency components from the two-dimensional series {\small $\bm{X}$} generated by the pre-filtering step. Theoretically, the functions encoded in {\small $\bm{\mathcal{L}}$} and {\small $\bm{\mathcal{H}}$} possess compact support, orthogonality, and regularity, making GHM particularly well-suited for multiscale analysis of complex temporal signals~\cite{geronimo1994fractal}.

{\vspace{1.2ex} \bf \em Remark:} Although the preset GHM parameters offer strong theoretical guarantees and desirable mathematical properties, their fixed nature limits flexibility when handling complex or heterogeneous data signals. In contrast, neural network-based parameter learning can adaptively adjust model parameters to better fit the data. To leverage the strengths of both approaches, we propose a novel architecture, \ie \model, that approximates the multi-wavelet decomposition process within a neural network framework. By learning transformation coefficients directly from the input series, \model captures both time-domain patterns and frequency-domain components in an end-to-end trainable manner, enabling more adaptive time series analysis.

\begin{figure}[t!]
    \centering
    \includegraphics[width=0.98\columnwidth]{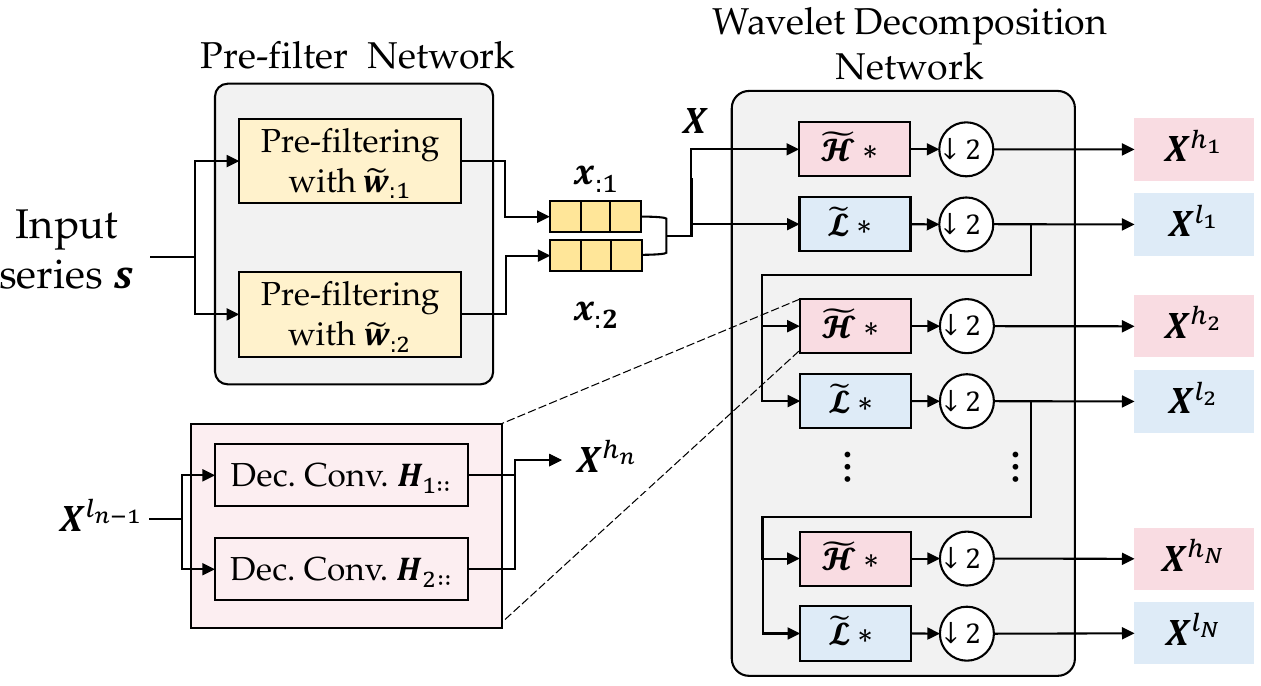}\vspace{-3mm}
    \caption{Illustration of the \model framework. Dec: Decomposition. Conv: Convolution. $\downarrow 2$ denotes the downsampling operation with a sample rate of 2. $*$ is the convolutional operator.}\label{fig:framework}\vspace{-4mm}
\end{figure}

The \model consists of two core modules: a Pre-filter Network, corresponding to the pre-filtering step of multi-wavelet decomposition, and a Wavelet Decomposition Network, which mirrors the wavelet decomposition step. \fig~\ref{fig:framework} illustrates the overall framework of \model. In the following sections, we detail these modules.

\subsection{Pre-filter Network of \model}\label{sec:pre_module}

The Pre-filter Network is designed to neuralize the pre-filtering phase of \mwd, enabling the model to adaptively transform the input time series into a multidimensional feature series. Each dimension corresponds to a distinct temporal pattern, allowing the network to capture diverse structural variations embedded within the original signal.

Since the pre-filtering process resembles a convolution operation, we neuralize it using a convolutional neural network, incorporating the \mwd kernel matrix {\small $\bm{P}$} defined in Eq.~\eqref{eq:qmatrix}. Specifically, given an input time series {\small $\bm{s} = (s_1, \ldots, s_t, \ldots, s_T)$} and a convolution kernel {\small $\bm{p}_{:1} = (p_{11}, \ldots, p_{k1}, \ldots, p_{K1})$} (which corresponds to a row vector in the GHM projection matrix {\small $\bm{P}$}), we define the convolution operation to compute the output series {\small $\bm{x} = (x_1, \ldots, x_t, \ldots, x_T)$} as:
\begin{equation}\label{eq:inner:conv1dBig}\small
	\bm{x} = \bm{p}_{:1} \ast \bm{s},
\end{equation}
where $\ast$ denotes the convolution operator. Each element $x_t$ in the output series $\bm{x}$ is computed as:
\begin{equation}\label{eq:inner:conv1d}\small
	x_t = \sum_{k=1}^{K} p_{k1} \cdot s_{t+k}.
\end{equation}

Recall the pre-filtering procedure in Eq.~\eqref{eq:prefilterequation}. The GHM-based multi-wavelet pre-filtering step can be reformulated in convolutional form as:
\begin{equation}\small
\begin{pmatrix}
    \bm{x}_{:1}\\
    \bm{x}_{:2}
\end{pmatrix} =
\begin{pmatrix}
    \bm{p}_{:1} \ast \bm{s} \\
    \bm{p}_{:2} \ast \bm{s}
\end{pmatrix},
\end{equation}
where the convolution kernels $\bm{p}_{:1}$ and $\bm{p}_{:2}$ are the row vectors of the GHM pre-filter matrix $\bm{P}$ in Eq.~\eqref{eq:qmatrix}. This formulation closely mirrors the original pre-filtering step in GHM decomposition and naturally supports implementation via convolutional layers within neural networks.

To enhance adaptability and enable the network to better capture time-domain patterns specific to the input series $\bm{s}$, we introduce a learnable parameter matrix {\small $\bm{W}^p \in \mathbb{R}^{|\bm{P}|}$} and define the trainable kernel as:
\begin{equation}\label{eq:pre-filtering_para}\small
    \tilde{\bm{W}} = \bm{P} + \bm{W}^p,
\end{equation}
\ie each element is computed as $\tilde{w}_{ij} = w_{ij}^p + p_{ij}$. The output of the pre-filtering network in \model is then given by:
\begin{equation}\label{eq:pre-filtering_net}\small
    \bm{X} =
    \begin{pmatrix}
        \bm{x}_{:1}\\
        \bm{x}_{:2}
    \end{pmatrix} =
    \sigma
    \begin{pmatrix}
        \tilde{\bm{w}}_{:1} \ast \bm{s} \\
        \tilde{\bm{w}}_{:2} \ast \bm{s}
    \end{pmatrix},
\end{equation}
where $\sigma(\cdot)$ denotes the sigmoid activation function. This non-linear activation is introduced to capture complex dependencies within the time series. The design in Eq.~\eqref{eq:pre-filtering_para} and Eq.~\eqref{eq:pre-filtering_net} preserves the theoretical structure of the GHM formulation while introducing learnable flexibility, enabling \model to adapt effectively through end-to-end training.

\subsection{Wavelet Decomposition Network of \model}\label{sec:decomp_module}

In this module, we neuralize the decomposition process of \model, enabling the multivariate series $\bm{X}$ in Eq.~\eqref{eq:pre-filtering_net} to be transformed into distinct frequency components in a learnable and data-adaptive manner.

We define a 2-dimensional convolution operator for multivariate time series. Given an input series {\small $\bm{X} \in \mathbb{R}^{N \times T}$} with $N$ features and $T$ time steps, and a kernel matrix {\small $\bm{K} \in \mathbb{R}^{M \times N}$}, the 2-dimensional convolution operation $\ast$ produces an output series as:
\begin{equation}\label{eq:conv2delement_2}\small
\bm{y} = \bm{K} \ast \bm{X},
\end{equation}
where {\small $\bm{y} = (y_1, \ldots, y_T)$}. The $t$-th item $y_t$ is computed as:
\begin{equation}\label{eq:conv2delement}\small
y_t = \sum_{m=1}^{M} \bm{k}_{m:} \cdot \bm{x}_{:,t+m},
\end{equation}
where {\small $\bm{k}_{m:}$} is the $m$-th row of the kernel matrix {\small $\bm{K}$}, and {\small $\bm{x}_{:,t+m}$} denotes the $(t+m)$-th column vector of the input series {\small $\bm{X}$}.

Furthermore, we extend the convolution kernel into a third-order tensor {\small $\bm{\mathcal{K}} \in \mathbb{R}^{M \times N \times K}$}. The convolution then produces a multivariate output series:
\begin{equation}\label{eq:convolution}\small
\bm{Y} =
    \begin{pmatrix}
        \bm{y}_{1:} \\
        \vdots \\
        \bm{y}_{K:}
    \end{pmatrix}
    = \bm{\mathcal{K}} \ast \bm{X} =
    \begin{pmatrix}
        \bm{K}_{1::} \ast \bm{X} \\
        \vdots \\
        \bm{K}_{K::} \ast \bm{X}
    \end{pmatrix},
\end{equation}
where $\bm{K}_{k::}$ denotes the $k$-th horizontal slice of the tensor $\bm{\mathcal{K}}$, and each $\bm{y}_{k:}$ is the corresponding output feature series.

For the GHM algorithm, the wavelet decomposition in Eq.~\eqref{eq:multi_decomposition} can be equivalently expressed as a 2-dimensional convolution operation with tensor kernels, following the formulation in Eq.~\eqref{eq:convolution}, as follows:
\begin{equation}\small
    \begin{aligned}
    {\bm{X}}^h &=
    \begin{pmatrix}
    {\bm{x}}^h_{1:} \\
    {\bm{x}}^h_{2:}
    \end{pmatrix}
    = \bm{\mathcal{H}} \ast \bm{X} =
    \begin{pmatrix}
    \bm{H}_{1::} \ast \bm{X} \\
    \bm{H}_{2::} \ast \bm{X}
    \end{pmatrix}, \\
    {\bm{X}}^l &=
    \begin{pmatrix}
    {\bm{x}}^l_{1:} \\
    {\bm{x}}^l_{2:}
    \end{pmatrix}
    = \;\bm{\mathcal{L}} \ast \bm{X} =
    \begin{pmatrix}
    \;\bm{L}_{1::} \ast \bm{X} \\
    \;\bm{L}_{2::} \ast \bm{X}
    \end{pmatrix},
    \end{aligned}
\end{equation}
where the kernel tensors {\small $\bm{\mathcal{L}}$} and {\small $\bm{\mathcal{H}}$} are defined in Eq.~\eqref{eq:lmatrix} and Eq.~\eqref{eq:hmatrix}, respectively.

In \model, the fixed wavelet and scaling tensors {\small $\bm{\mathcal{H}}$} and {\small $\bm{\mathcal{L}}$} are enhanced with two learnable parameter tensors {\small $\bm{\mathcal{W}}^h \in \mathbb{R}^{|\bm{\mathcal{H}}|}$} and {\small $\bm{\mathcal{W}}^l \in \mathbb{R}^{|\bm{\mathcal{L}}|}$} as follows:
\begin{equation}\label{eq:decomposition_para}\small
    \tilde{\bm{\mathcal{H}}} = \bm{\mathcal{W}}^h + \bm{\mathcal{H}},\quad\quad \tilde{\bm{\mathcal{L}}} = \bm{\mathcal{W}}^l + \bm{\mathcal{L}}.
\end{equation}
This formulation allows \model to adaptively adjust the wavelet and scaling functions based on the characteristics of the input data. By augmenting the traditional GHM filters with learnable parameters, the model gains greater flexibility and adaptability, leading to improved frequency component extraction tailored to diverse time series.

Using the learnable parameters defined in Eq.~\eqref{eq:decomposition_para}, \model recursively extracts high- and low-frequency components of the input series as follows:
\begin{equation}\small
\begin{aligned}
\bm{X}^{h_n} &= \sigma \left( ( \tilde{\bm{\mathcal{H}}} \ast \bm{X}^{l_{n-1}} ) \downarrow 2 \right), \\
\bm{X}^{l_n} &= \sigma \left( ( \tilde{\bm{\mathcal{L}}} \ast \bm{X}^{l_{n-1}} ) \downarrow 2 \right),
\end{aligned}
\end{equation}
where $\sigma(\cdot)$ denotes the sigmoid activation function. The use of $\sigma(\cdot)$ enables the model to capture complex, non-linear relationships within the frequency domain, thereby enhancing its capacity to model intricate temporal patterns across multiple decomposition layers.

Similar to the multi-wavelet decomposition in Sec.~\ref{sec:multi-wavelet}, the output of \model with $N$ layers is a set of frequency components:
\begin{equation}\label{eq:outputs}\small
\bm{\mathcal{X}} = \left\{\bm{X}^{h_1}, \bm{X}^{h_2}, \ldots, \bm{X}^{h_N}, \bm{X}^{l_N}\right\},
\end{equation}
where {\small $\bm{X}^{h_1} \in \mathbb{R}^{2 \times (T / 2)}$} represents the highest-frequency component, and {\small $\bm{X}^{l_N} \in \mathbb{R}^{2 \times (T / 2^N)}$}  corresponds to the lowest-frequency component. This set {\small $\bm{\mathcal{X}}$} constitutes the final time-frequency features extracted by \model.

{\vspace{1.2ex} \bf \em Remark:} To sum up, \model uses a deep neural network framework to approximately implement the GHM multi-wavelet decomposition. Achieving the GHM algorithm within such a framework offers several advantages. First, the parameters in GHM can be fine-tuned via backpropagation with task-specific loss functions, enabling the extracted frequency features to carry more relevant information than those obtained from the traditional GHM method. Second, the output series produced by \model can be fed into downstream neural networks for further analysis, facilitating end-to-end training of the entire pipeline (see Sec.~\ref{sec:ensemble}) and thereby improving overall model performance.




\subsection{Orthogonality Regularization}

In this part, we first introduce the orthogonality property of \mwd, and then design an orthogonality regularization term to constrain the convolution kernels of \model to align with the orthogonality property of \mwd.

\mwd possesses properties such as orthogonality, tight support, and regularity. Among these, orthogonality serves as the foundation, ensuring that the input series can be decomposed and reconstructed completely, stably, and without redundancy. The other properties build upon this foundation to further enhance performance and facilitate practical applications. Therefore, this paper focuses on preserving the most critical property of \mwd: orthogonality. Its detailed introduction is given in Sec. 5 of the Supplementary Materials (SM).

To satisfy the orthogonality requirement in the proposed \model framework, we must ensure that the corresponding convolutional kernels {\small $\tilde{\bm{\mathcal{H}}}$} and {\small $\tilde{\bm{\mathcal{L}}}$} are as orthonormal as possible. Specifically, given {\small $\tilde{\bm{H}}_{k::}$} and {\small $\tilde{\bm{L}}_{k::}$}, which denote the $k$-th horizontal slices of the tensors {\small $\tilde{\bm{\mathcal{H}}}$} and {\small $\tilde{\bm{\mathcal{L}}}$} respectively, we expect that
\begin{equation}\label{eq:orth_kernel}\scriptsize
\begin{aligned}
   \tilde{\bm{H}}_{1::}\cdot \tilde{\bm{H}}_{1::}^\top = \tilde{\bm{H}}_{2::}\cdot \tilde{\bm{H}}_{2::}^\top = \bm{I},~~ &
   \tilde{\bm{H}}_{1::}\cdot \tilde{\bm{H}}_{2::}^\top = \tilde{\bm{H}}_{2::}\cdot \tilde{\bm{H}}_{1::}^\top = \bm{0},\\
   \tilde{\bm{L}}_{1::}\cdot \tilde{\bm{L}}_{1::}^\top = \tilde{\bm{L}}_{2::}\cdot \tilde{\bm{L}}_{2::}^\top = \bm{I},~~ &
   \tilde{\bm{L}}_{1::}\cdot \tilde{\bm{L}}_{2::}^\top = \tilde{\bm{L}}_{2::}\cdot \tilde{\bm{L}}_{1::}^\top = \bm{0},\\
   \tilde{\bm{H}}_{1::}\cdot \tilde{\bm{L}}_{1::}^\top = \tilde{\bm{H}}_{2::}\cdot \tilde{\bm{L}}_{2::}^\top = \bm{0},~~ &
   \tilde{\bm{H}}_{1::}\cdot \tilde{\bm{L}}_{2::}^\top = \tilde{\bm{H}}_{2::}\cdot \tilde{\bm{L}}_{1::}^\top = \bm{0},\\
   \tilde{\bm{L}}_{1::}\cdot \tilde{\bm{H}}_{1::}^\top = \tilde{\bm{L}}_{2::}\cdot \tilde{\bm{H}}_{2::}^\top = \bm{I},~~ &
   \tilde{\bm{L}}_{1::}\cdot \tilde{\bm{H}}_{2::}^\top = \tilde{\bm{L}}_{2::}\cdot \tilde{\bm{H}}_{1::}^\top = \bm{0},
\end{aligned}
\end{equation}
where $\bm{I}$ is the identity matrix and $\bm{0}$ is a zero matrix.
\equ~\eqref{eq:orth_kernel} indicates that: $i)$ each convolutional kernel forms a row-orthonormal matrix, and $ii)$ each convolutional kernel is orthogonal to every other convolutional kernel.

Each condition in \equ~\eqref{eq:orth_kernel} is supposed to be satisfied. However, directly optimizing all these terms would introduce numerous optimization objectives, leading to a computationally expensive and inefficient process, and potentially causing unstable training.
To address this, this paper adopts the block-Toeplitz (BT) method~\cite{bottcher2012introduction} to efficiently represent and handle large numbers of 2D convolution kernels.
Specifically, a Toeplitz matrix is one where elements are constant along each diagonal. A Block-Toeplitz matrix generalizes this concept: instead of scalars, the elements are small matrices (blocks), and these blocks remain constant along each diagonal, similar to the structure of a standard Toeplitz matrix.
Given the convolution kernels of \model, they can be organized into a BT matrix {\small $\bm{B}$} as shown below:
\begin{equation}\scriptsize
\bm{B}^\top =
\begin{pmatrix}
    \tilde{\bm{H}}_{1::} & 0 & \tilde{\bm{H}}_{2::} & 0 & \tilde{\bm{L}}_{1::} & 0 & \tilde{\bm{L}}_{2::} & 0\\
    0 &  \tilde{\bm{H}}_{1::} & 0 & \tilde{\bm{H}}_{2::} & 0 &  \tilde{\bm{L}}_{1::} & 0 & \tilde{\bm{L}}_{2::}\\
\end{pmatrix},
\end{equation}
Based on {\small $\bm{B}$}, the optimization objective for kernel orthogonality in \equ~\eqref{eq:orth_kernel} can be reformulated as
\begin{equation}\small\label{eq:orthogonal regularization}
\mathcal{L}_{o}  = \Vert\bm{B} \cdot \bm{B}^{\top}-\bm{I} \Vert_{F}^2,
\end{equation}
where $\Vert\cdot\Vert_F$ denotes the Frobenius norm. With the BT method, the orthogonality regularization is formulated as a single matrix multiplication operation. This allows deep learning libraries to leverage parallelization to accelerate computation, thereby ensuring computational efficiency.

{\vspace{1.1ex} \bf \em Remark:} The orthogonality regularization in this part offers several advantages.
First, when filters (kernels) are learned to be as orthogonal as possible, they become decorrelated, resulting in filter responses that are significantly less redundant.
Moreover, the orthonormal kernel matrix helps gradients back-propagate stably, thus preventing gradient explosion and gradient vanishing. This is because multiplying a matrix by an orthonormal matrix preserves its norm.

\section{\model-based Time Series Analysis}\label{sec:ensemble}

In this section, we propose two \model-based network architectures to leverage the capability of \model for different downstream tasks, including time series classification (TSC) and time series forecasting (TSF).


\subsection{\model-based Time Series Classification}

The time series classification (TSC) task aims to predict the category label of a time series based on its temporal features. For instance, diagnosing whether a patient has heart disease can be formulated as a classification problem using electrocardiogram (ECG) signal sequences as input. A key challenge in TSC lies in effectively extracting discriminative features from the input time series. \model provides a powerful solution by jointly capturing both temporal structures and frequency-domain characteristics of time series through its time-frequency decomposition framework. Building on this capability, we propose a Time-Frequency Boosting Classification (\cls) network, which exploits the rich representations in the frequency components extracted by \model layer by layer. To further enhance the discriminative power of the frequency features, we introduce a Frequency Contrastive Learning (\fcl) loss, enabling the network to be trained in a pre-training followed by fine-tuning paradigm. This design encourages the model to learn generalized and class-discriminative representations that are robust across diverse TSC scenarios.

\subsubsection{Time-Frequency Boosting Classification}

\begin{figure}[t]
    \centering
    \includegraphics[width=0.9\columnwidth]{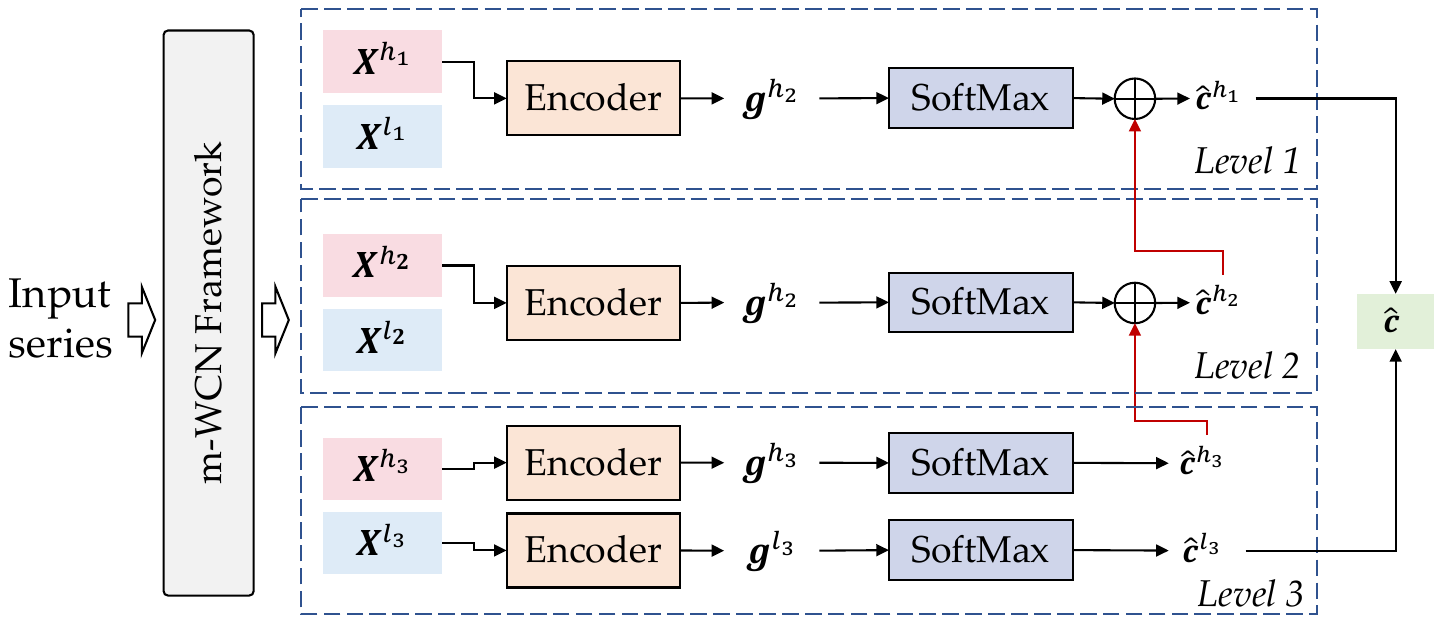}\vspace{-3mm}
    \caption{The \cls network for classification: A three-level example.}
    \label{fig:RCF Framework}\vspace{-4mm}
\end{figure}

In \model, the extracted frequency components represent complementary aspects of the input signal. While each component carries a certain degree of discriminative power, the low-frequency components typically encode more informative features for classification tasks than high-frequency components in wavelet decompositions~\cite{wang2018multilevel}. Motivated by this insight, the \cls network adopts a boosting strategy to exploit the hierarchical relationship between low- and high-frequency components. The classification process begins with the lowest-frequency component to generate an initial prediction. Higher-frequency components are then introduced sequentially to refine this prediction by learning to correct the residual errors from earlier stages. This hierarchical boosting mechanism enables the model to integrate both stable global patterns and subtle local variations in the signal. \fig~\ref{fig:RCF Framework} illustrates the architecture of \cls.

Given the lowest-frequency component in Eq.~\eqref{eq:outputs}, i.e., {\small $\bm{X}^{l_N}$}, the \cls network uses an encoder $\mathrm{Enc}(\cdot)$ to transform it into a representation vector:
\begin{equation}\label{eq:encoder_N}\small
\bm{g}^{l_N} = \mathrm{Enc}\left(\bm{X}^{l_N}; \bm{\Theta}^{l_N}_e\right),
\end{equation}
where $\bm{\Theta}^N_e$ denotes the trainable parameters of the encoder, and {\small $\bm{g}^{l_N}$} is the resulting representation vector. We implement $\mathrm{Enc}(\cdot)$ using TimesNet~\cite{wu2023timesnet}, due to its effectiveness in feature extraction and transformation.

Subsequently, the \cls network applies a Softmax classifier to predict the class label from {\small $\bm{g}^{l_N}$}:
\begin{equation}\label{eq:decoder_N}\small
\hat{\bm{c}}^{l_N} = \mathrm{SoftMax}\left(\bm{g}^{l_N}\right),
\end{equation}
where {\small $\hat{\bm{c}}^{l_N}$} is the probability vector representing the prediction for the one-hot class label.
The operations defined in \equ~\eqref{eq:encoder_N} and \equ~\eqref{eq:decoder_N} together constitute the $N$-th layer of the \cls network.


For the $n$-th high-frequency components, where {\small $n \in \{0, \ldots, N-1\}$}, the \cls network applies an encoder to generate the representation vector for the high-frequency component {\small $\bm{X}^{h_n}$} extracted from the $n$-th decomposition layer of \model:
\begin{equation}\label{eq:encoder_n}\small
\bm{g}^{h_n} = \mathrm{Enc}\left(\bm{X}^{h_n}; \bm{\Theta}^{h_n}_e\right),
\end{equation}
Then, the class label prediction is generated using a residual formulation:
\begin{equation}\label{eq:decoder_n}\small
\hat{\bm{c}}^{h_n} = \hat{\bm{c}}^{h_{n+1}} + \mathrm{SoftMax}\left(\bm{g}^{h_n}\right).
\end{equation}
Intuitively, Eq.~\eqref{eq:decoder_n} uses {\small $\bm{g}^{h_n}$} to estimate the residual error between the current prediction and the one from the higher layer, \ie {\small $\hat{\bm{c}}^{h_n} - \hat{\bm{c}}^{h_{n+1}} = \mathrm{SoftMax}\left(\bm{g}^{h_n}\right)$}. In this way, the representation vector {\small $\bm{g}^{h_n}$} learns to compensate for what the $(n+1)$-th layer fails to model, forming a boosting-style learning mechanism that progressively refines the prediction across layers.

Finally, the final class label prediction is given by the first layer. According to the recursive expression in Eq.~\eqref{eq:decoder_n}, the final label prediction is calculated as an ensemble form of the predictions based on the lowest frequency component and high-frequency components, \ie
\begin{equation}\label{eq:final}\small
        \hat{\bm{c}}  
        = \frac{1}{N+1}\left(\mathrm{SoftMax}\left(\bm{g}^{l_N}\right) + \sum_{n=1}^{N} \mathrm{SoftMax}\left(\bm{g}^{h_n}\right)\right).
\end{equation}

{\vspace{1.2ex} \bf \em Remark:} In the \cls model, the frequency components extracted from all levels of \model are utilized to generate an ensemble class prediction. As these components reflect different frequency resolutions~\cite{mallat1989theory}, \cls effectively captures diverse perspectives of the input time series by aggregating information across multiple spectral scales. This design enables \cls to operate as a multi-view learning framework, enhancing its ability to achieve high-performance time series classification. Moreover, the classifier at the $n$-th level refines its prediction $\hat{\bm{c}}^n$ based on both its own representation $\bm{g}^n$ and the output from the $(n+1)$-th level, $\hat{\bm{c}}^{h_{n+1}}$. This residual formulation allows the model to incrementally integrate knowledge that is not captured by higher-frequency components alone, thereby improving the overall classification accuracy through progressive refinement.


\subsubsection{Frequency Contrastive Learning}

In the \cls model, each frequency component is encoded into a representation vector $\bm{g}^n$. To enhance the discriminative capacity of these representations, we introduce a self-supervised Frequency Contrastive Learning (FCL) strategy to pre-train the parameters of \cls. This approach encourages the model to learn semantically meaningful frequency-related embeddings.


In \cls, the representations $\bm{g}^n$ obtained at different decomposition levels describe the same input signal from multiple frequency perspectives. These multi-level representations can thus be regarded as natural augmentations of the original signal. Based on this intuition, we treat pairs of representations $\bm{g}^n$ and $\bm{g}^m$ extracted from the same time series instance as positive samples, while representations from different time series are treated as negative samples.


Specifically, for an input time series {\small $\bm{s}_i$}, we obtain $N{+}1$ representation vectors from different frequency components, denoted as {\small $\bm{g}_i^\pi$}, where {\small $\pi \in \Pi = \{h_1, \ldots, h_N, l_N\}$} denotes the index of a frequency component. To facilitate contrastive learning, a nonlinear projection head $g$ is used to map each level-specific representation into a latent space: {\small $\bm{v}_i^\pi = g(\bm{g}_i^\pi)$}. In this latent space, we define pairwise contrastive objectives using the InfoNCE loss. For two representations {\small $\bm{v}_i^{\pi_1}$} and {\small $\bm{v}_i^{\pi_2}$} derived from different frequency levels of the same input series sample {\small $\bm{S}_i$}, the contrastive loss is computed as:
\begin{equation}\small
\begin{split}
\mathcal{L}_i(\pi_1, \pi_2) &= -\log \frac{h\left(\bm{v}_i^{\pi_1}, \bm{v}_i^{\pi_2}\right)}{h\left(\bm{v}_i^{\pi_1}, \bm{v}_i^{\pi_2}\right) + \mathrm{NP}}, \\
\text{where} \quad \mathrm{NP} &= \sum_{j \neq i} h\left(\bm{v}_i^{\pi_1}, \bm{v}_j^{\pi_1}\right) + \sum_{j \neq i} h\left(\bm{v}_i^{\pi_2}, \bm{v}_j^{\pi_2}\right).
\end{split}
\end{equation}
Here, $h(\bm{v}_a, \bm{v}_b) = \exp(\mathrm{sim}(\bm{v}_a, \bm{v}_b) / \tau)$, where $\mathrm{sim}(\cdot, \cdot)$ denotes cosine similarity and $\tau$ is a temperature parameter controlling the sharpness of similarity scores. The projection head $g$ is implemented as a two-layer MLP.

The final frequency contrastive loss across a batch of $I$ samples is defined as:
\begin{equation}\label{eq:contrastive learning}\small
\mathcal{L}_c = \frac{1}{H} \sum_{i=1}^{I} \sum_{\pi_1 \in \Pi} \sum_{\pi_2 \neq \pi_1} \mathcal{L}_i(\pi_1, \pi_2),
\end{equation}
where {\small $\Pi = \{h_1, \ldots, h_N, l_N\}$}, and {\small $H = I \times N \times (N+1)$} is the total number of contrastive pairs in the batch. This framework encourages the model to pull together representations of the same input across different frequency components while pushing apart representations from different inputs. As a result, the model learns semantically rich, frequency-related embeddings that provide strong initialization for downstream time series classification tasks via fine-tuning.


\subsubsection{Optimization for \cls}

For the final classification loss of \cls, we adopt a layer-wise supervision strategy to enhance training effectiveness. Specifically, given a set of $I$ input time series samples, we define {\small $\hat{\bm{c}}_i^\pi$} as the predicted class label generated by Eq.~\eqref{eq:decoder_n} for the $i$-th sample, where the superscript $\pi$ denotes the corresponding frequency component (either $h_n$ or $l_N$). The cross-entropy loss for {\small $\hat{\bm{c}}_i^\pi$} is defined as:
\begin{equation}\label{eq:loss1}\small
\mathcal{\tilde{L}}_{e}(\pi) = - \frac{1}{I} \sum_{i=1}^{I} \left( \bm{c}_i^\top \ln \hat{\bm{c}}_i^\pi + \left(1 - \bm{c}_i\right)^\top \ln\left(1 - \hat{\bm{c}}_i^\pi\right) \right),
\end{equation}
where {\small $\bm{c}_i$} denotes the one-hot encoded ground truth label for the $i$-th sample.

For an \cls model with $N$ \model frequency decomposition layers, the overall cross-entropy objective is defined as a weighted sum of all individual layer losses {\small $\tilde{\mathcal{L}}_{e}(\pi)$}:
\begin{equation}\label{eq:J_rcf}\small
\mathcal{L}_e = \sum_{n=1}^{N} \frac{N-n+1}{N} \mathcal{\tilde{L}}_{e}(h_n) + \frac{1}{N}\mathcal{\tilde{L}}_{e}(l_N).
\end{equation}
This weighting scheme ensures that the final prediction {\small $\hat{\bm{c}} = \hat{\bm{c}}^{h_1}$} — which integrates information from all frequency components — receives the highest weight of $1$, while the prediction  {\small $\hat{\bm{c}}^{l_N}$} — based solely on the lowest-frequency component — receives the smallest weight of {\small $1/N$}. This design prioritizes the supervisory signal for the most comprehensive prediction while still guiding the learning process at earlier layers.


The overall objective for TSC is derived by integrating the classification loss in Eq.~\eqref{eq:J_rcf}, the contrastive loss in Eq.~\eqref{eq:contrastive learning}, and the orthogonal regularization in Eq.~\eqref{eq:orthogonal regularization} as
\begin{equation}\label{eq:total loss for TSC}\small
\mathcal{L}_{\mathrm{TSC}} = \mathcal{L}_e + \gamma_1 \mathcal{L}_{c} + \gamma_2 \mathcal{L}_{o},
\end{equation}
where $\gamma_1$ and $\gamma_2$ are hyperparameters that balance the contributions of different losses. 

\begin{figure}[t]
    \centering
    \includegraphics[width=0.9\columnwidth]{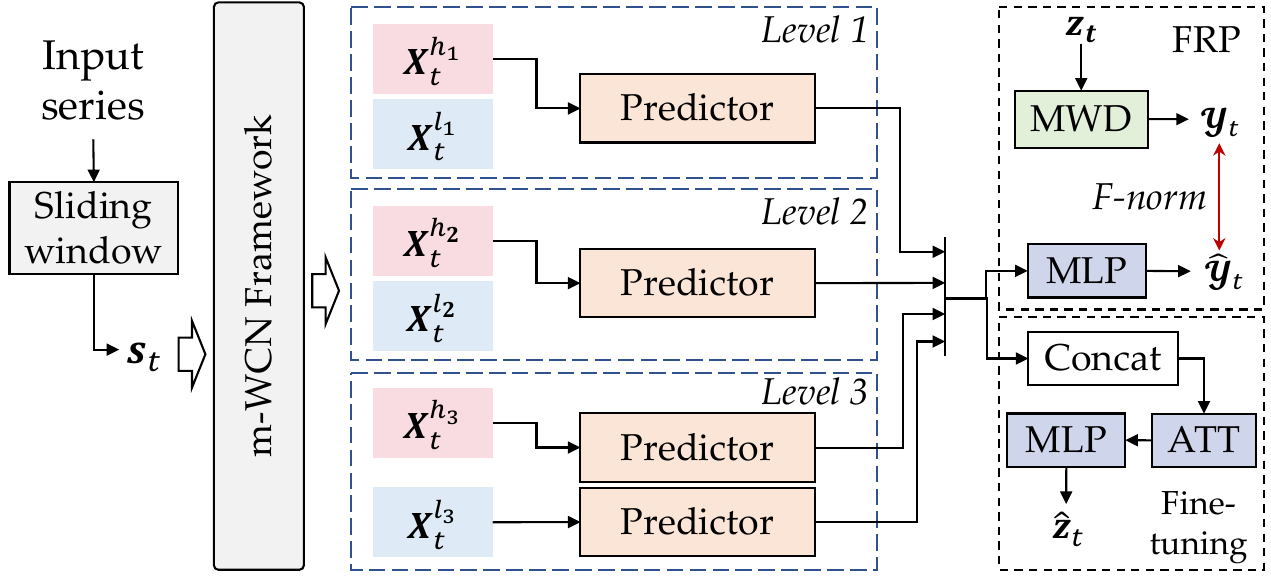}\vspace{-3mm}
    \caption{The \ftb model for forecasting: A three-level example. FRP: Frequency Representation Pre-training.}
    \label{fig:MWDCN-lstm}\vspace{-4mm}
\end{figure}

\subsection{\model-based Time Series Forecasting}

The time series forecasting (TSF) task aims to predict future values of a time series based on its historical observations, typically following an autoregressive paradigm. Unlike time series classification (TSC), where class labels may be associated with any frequency component, TSF tasks often exhibit frequency-aligned dependencies — meaning that each frequency component in the future series tends to correlate most strongly with the corresponding frequency component in the historical series. Motivated by this insight, we propose a \model-based Frequency {TSMixer} Bagging framework, abbreviated as \ftb, for time series forecasting. To support training of \ftb, we further introduce a Frequency Representation Pre-training (FRP) strategy to enhance frequency-aware learning. The overall architecture of the proposed forecasting model is illustrated in \fig~\ref{fig:MWDCN-lstm}.


\subsubsection{Bagging of TSMixer for Forecasting}

In the \ftb network, we decompose a complex TSF task into multiple sub-problems, each corresponding to forecasting a specific frequency component extracted by \model. Since each frequency component captures a relatively simpler temporal pattern — either long-term trends or short-term fluctuations — these sub-problems are easier to model individually. To handle them effectively, \ftb employs separate {{TSMixer-based}} predictors for each frequency component. The predictions from these parallel sub-models are then aggregated to produce the final forecasting result, enabling the model to capture both coarse-grained and fine-grained temporal dynamics in a frequency-aware manner.

Specifically, given an input time series of infinite length, we apply a sliding window of size $T+1$ over the past to the current time step $t$, resulting in the input segment:
\begin{equation}\small
\bm{s}_t = \left(s_{t-T}, \ldots, s_{t-1}, s_t\right).
\end{equation}
In the TSF setting, the goal is to predict the future segment of the series starting from time $t{+}1$ over a prediction horizon of length $L$, defined as:
\begin{equation}\label{eq:output}\small
\bm{z}_{t} = \left(s_{t+1}, \ldots, s_{t+l}, \ldots, s_{t+L}\right).
\end{equation}

The \ftb model first applies \model to decompose the input series $\bm{s}_t$ into a set of frequency components: {\small ${\bm{X}_t^{h_1}, \ldots, \bm{X}_t^{h_N}, \bm{X}_t^{l_N}}$}, capturing both high- and low-frequency patterns. Each component is then processed independently using a {{TSMixer}} network to obtain its frequency-specific representation sequence:
\begin{equation}\label{eq:frequency_representation}\small
    \begin{aligned}
        \bm{E}_t^{h_n} &= \left(\bm{e}^{h_n}_{t - \frac{T}{2^{n+1}}}, \ldots, \bm{e}^{h_n}_t\right) = \mathrm{TF}\left(\bm{X}_t^{h_n}; \bm{\Theta}^{h_n}\right), \\
        \bm{E}_t^{l_N} &= \left(\bm{e}^{l_N}_{t - \frac{T}{2^{N+1}}}, \ldots, \bm{e}^{l_N}_t\right) = \mathrm{TF}\left(\bm{X}_t^{l_N}; \bm{\Theta}^{l_N}\right),
    \end{aligned}
\end{equation}
where $\mathrm{TF}(\cdot)$ denotes the {{TSMixer encoder}}, and $\bm{\Theta}^{h_n}$ and $\bm{\Theta}^{l_N}$ are the learnable parameters for the $n$-th high-frequency and final low-frequency predictors, respectively. 

Then, the \ftb model applies an attention mechanism to integrate the representation sequences from all frequency components:
\begin{equation}\small
    \bm{U} = \mathrm{ATT}\left(\bm{E}_t^{h_1} \| \ldots \| \bm{E}_t^{h_N} \| \bm{E}_t^{l_N}\right),
\end{equation}
where $\mathrm{ATT}(\cdot)$ denotes the attention network and $\|$ indicates concatenation along the temporal dimension. This step captures the inter-frequency dependencies and synthesizes a unified context representation $\bm{U}$.

Finally, a multi-layer perceptron (MLP) is used to generate the forecasted future sequence:
\begin{equation}\label{eq:tsf_output}\small
\hat{\bm{z}}_t = \mathrm{MLP}(\bm{U}),
\end{equation}
where $\hat{\bm{z}}_t = (\hat{s}_{t+1}, \ldots, \hat{s}_{t+l}, \ldots, \hat{s}_{t+L})$ is the predicted future segment corresponding to $\bm{z}_{t+1}$. Each $\hat{s}_{t+l}$ represents the predicted value at time step $t+l$, completing the end-to-end forecasting pipeline.

{\vspace{1.2ex} \bf \em Remark:} In the TSF task, different frequency components reflect temporal dynamics at varying scales, \ie low frequency captures long-term trends, while high frequency encodes short-term variations. All these components are equally essential for accurate forecasting. To effectively utilize this multi-scale information, the \ftb model adopts a bagging approach: each frequency-specific representation independently contributes to the prediction of the future series. This design enhances the model's ability to capture rich temporal dependencies across all spectral bands.

\subsubsection{Frequency Representation Pre-training}

To improve the quality and forecasting relevance of the frequency-specific representations in Eq.~\eqref{eq:frequency_representation}, we propose a Frequency Representation Pre-training (FRP) strategy. This approach aims to enhance each frequency component's ability to capture predictive information by encouraging its representation to reconstruct the corresponding component in the future series.

Specifically, given the future time series segment $\bm{z}_t$ defined in Eq.~\eqref{eq:output}, we apply GHM multi-wavelet decomposition to extract its frequency components: {\small $\mathcal{Y}_t = \{\bm{Y}_t^{h_1}, \ldots, \bm{Y}_t^{h_N}, \bm{Y}_t^{l_N}\}$}. Each component $\bm{Y}_t^{h_n}$ serves as the prediction label for pre-training the representation $\bm{E}_t^{h_n}$ generated in Eq.~\eqref{eq:frequency_representation}. Next, we employ an MLP network to predict each future frequency component from its corresponding past representation:
\begin{equation}\small
\hat{\bm{Y}}_t^{h_n} = \mathrm{MLP}\left(\bm{E}_t^{h_n}; \bm{\Theta}_p^{h_n}\right),
\end{equation}
where $\bm{\Theta}_p^{h_n}$ are learnable parameters for the $n$-th predictor.

The pre-training loss for time step $t$ is formulated as:
\begin{equation}\label{eq:loss_pre_TSF}\small
\mathcal{L}_t = \sum_{n=1}^N \frac{1}{D^n} \left\Vert \bm{Y}_t^{h_n} - \hat{\bm{Y}}_t^{h_n} \right\Vert_F^2 + \frac{1}{D^N} \left\Vert \bm{Y}_t^{l_N} - \hat{\bm{Y}}_t^{l_N} \right\Vert_F^2,
\end{equation}
where $D^n$ is the dimensionality of the $n$-th frequency component, used to normalize losses across different scales.

Over $M$ forecast training samples, the total pre-training objective is defined as:
\begin{equation}\small
\mathcal{L}_p = \sum_{t=1}^{M} \mathcal{L}_t + \gamma_3 \mathcal{L}_{o},
\end{equation}
where $\gamma_3$ is a hyperparameter and $\mathcal{L}_{o}$ denotes the orthogonality regularization term introduced in Eq.~\eqref{eq:orthogonal regularization}.

\subsubsection{Optimization for \ftb}

In the fine-tuning stage, the \ftb model is initialized with parameters learned during the pre-training phase. The entire model is then trained in an end-to-end manner using the mean square error (MSE) loss to minimize the discrepancy between the predicted and ground-truth future sequences. Specifically, given $M$ time series forecasting samples, the fine-tuning objective is defined as:
\begin{equation}\label{eq:J_mlstm}\small
\mathcal{L}_{\mathrm{TSF}} = \frac{1}{M}  \sum_{t=1}^M \left\Vert \bm{z}_t - \hat{\bm{z}}_t \right\Vert_2^2,
\end{equation}
where $\hat{\bm{z}}_t$ is the predicted sequence generated by Eq.~\eqref{eq:tsf_output}, and $\bm{z}_t$ is the ground truth future segment. 

\subsection{Complexity Discussion}

\paratitle{Complexity Discussion.} Let $T$ denote the input length and $N$ denote the decomposition depth. The additional computational cost introduced by m-WCN mainly comes from two modules: the pre-filter network and the wavelet decomposition network. In our GHM-based implementation, both the convolutional kernel sizes and the number of channels are fixed constants. Therefore, the complexity of m-WCN grows linearly with the input length.
\begin{itemize}[leftmargin=*]
    \item {\em Time complexity.} The pre-filter network scans the input sequence once using fixed-size convolutional filters, leading to a time complexity of $O(T)$. For the wavelet decomposition network, the sequence length is downsampled by a factor of 2 after each decomposition layer. Thus, the input length to the $n$-th decomposition layer is $T/2^{n-1}$. Since each layer applies fixed-size convolutional filters with a fixed number of channels, the total computational cost over all $N$ decomposition layers is $\sum_{n=1}^{N} O(T/2^{n-1}) = O(T)$. Therefore, the overall additional time complexity introduced by m-WCN is linear in $T$.

    \item {\em Space complexity.} The space complexity is also $O(T)$. The final retained frequency components consist of the high-frequency components from all decomposition layers and the final low-frequency component. Their total sequence length is bounded by $\sum_{n=1}^{N} T/2^n + T/2^N \leq T$, up to constant factors determined by the fixed channel number. During training, storing intermediate activations for back-propagation also leads to a geometric sum over sequence lengths and therefore remains linear in $T$.
\end{itemize}

In summary, m-WCN introduces only linear additional time and memory overhead with respect to the input length. This indicates that the proposed decomposition module is computationally efficient and can be integrated into downstream classification and forecasting models without causing excessive complexity.


\begin{table*}[!t]
  \centering
  \caption{Classification performance comparison on 64 UCR time series datasets regarding error rate. The Dataset column is in the form of ``Abbr/Type ID'', where the dataset's full name and corresponding type are in Sec. 1.1 of SM. The best result is bold, while the second-best result is underlined.}\vspace{-2mm}
  \resizebox{\linewidth}{!}{
    \setlength{\tabcolsep}{0.5mm}
    \renewcommand{\arraystretch}{1.1}
\begin{tabular}{r|cccccccc||r|cccccccc}
\toprule
\textbf{Dataset} & miniRocket & Hivecote2 & TS-Chief & OS-CNN & TimeURL & {MILLET} & mWDN  & \cls & \textbf{Dataset} & miniRocket & Hivecote2 & TS-Chief & OS-CNN & TimeURL & {MILLET} & mWDN  & {\cls} \\
\midrule
Comput/1 & 0.268  & \underline{0.240} & 0.300  & 0.293  & 0.264  & \underline{0.240} & 0.360  & \textbf{0.220} & CrkX/4 & 0.179  & 0.172  & 0.179  & \textbf{0.145} & 0.267  & \underline{0.154} & 0.216  & 0.168  \\
ElecD/1 & \underline{0.258} & 0.274  & 0.284  & 0.276  & 0.273  & 0.271  & 0.342  & \textbf{0.235} & CrkY/4 & 0.172  & 0.154  & 0.200  & \textbf{0.133} & 0.264  & 0.149  & 0.172  & \underline{0.141} \\
LrgKA/1 & 0.125  & \underline{0.080} & 0.232  & 0.104  & 0.087  & 0.096  & 0.152  & \textbf{0.067} & CrkZ/4 & 0.172  & 0.141  & 0.169  & \underline{0.137} & 0.254  & \textbf{0.136} & 0.162  & 0.169  \\
RefrD/1 & 0.520  & 0.448  & 0.429  & 0.497  & \textbf{0.379} & 0.491  & 0.493  & \underline{0.424} & Haptics/4 & 0.471  & \underline{0.445} & 0.471  & 0.490  & 0.451  & 0.513  & 0.461  & \textbf{0.427} \\
ScrnT/1 & 0.539  & \underline{0.429} & 0.501  & 0.474  & 0.467  & \textbf{0.405} & 0.480  & 0.461  & ISkat/4 & 0.524  & \underline{0.456} & 0.473  & 0.571  & 0.556  & 0.533  & 0.566  & \textbf{0.445} \\
SmKA/1 & 0.173  & \textbf{0.163} & 0.176  & 0.279  & 0.233  & 0.227  & 0.221  & \underline{0.168} & ToeS1/4 & 0.039  & 0.035  & \underline{0.031} & 0.046  & 0.066  & 0.039  & \underline{0.031} & \textbf{0.026} \\
\cline{1-9} ECG200/2 & 0.090  & 0.140  & 0.160  & 0.092  & \underline{0.070} & 0.100  & \underline{0.070} & \textbf{0.050} & ToeS2/4 & 0.077  & 0.062  & \underline{0.038} & 0.054  & 0.131  & 0.069  & 0.138  & \underline{0.038} \\
ECG5000/2 & 0.055  & \underline{0.053} & 0.054  & 0.060  & 0.057  & 0.061  & 0.070  & \textbf{0.033} & UWAll/4 & 0.029  & \underline{0.025} & 0.030  & 0.058  & 0.039  & 0.046  & 0.028  & \textbf{0.020} \\
NFET1/2 & 0.051  & 0.050  & 0.083  & \underline{0.037} & 0.049  & 0.055  & \textbf{0.026} & 0.040  & UWX/4 & 0.152  & \textbf{0.142} & 0.157  & 0.178  & 0.192  & 0.184  & 0.218  & \underline{0.148} \\
NFET2/2 & 0.036  & \underline{0.034} & 0.052  & 0.040  & 0.046  & 0.042  & \textbf{0.028} & 0.044  & UWY/4 & 0.224  & \underline{0.219} & 0.229  & 0.243  & 0.257  & 0.250  & 0.232  & \textbf{0.210} \\
\cline{1-9}Adiac/3 & 0.184  & 0.194  & 0.202  & 0.165  & 0.182  & 0.174  & \underline{0.155} & \textbf{0.143} & UWZ/4 & \underline{0.199} & 0.201  & 0.214  & 0.236  & 0.247  & 0.253  & 0.265  & \textbf{0.164} \\
Arrow/3 & 0.137  & 0.131  & 0.194  & 0.162  & \underline{0.103} & 0.200  & 0.181  & \textbf{0.091} & Worms/4 & 0.260  & 0.260  & \underline{0.182} & 0.235  & \underline{0.182} & 0.208  & 0.195  & \textbf{0.169} \\
BChic/3 & 0.100  & 0.100  & \underline{0.050} & 0.115  & 0.100  & \underline{0.050} & 0.100  & \textbf{0} & WormT/4 & 0.221  & 0.195  & \underline{0.169} & 0.343  & \underline{0.169} & 0.286  & 0.208  & \textbf{0.130} \\
\cline{10-18}DPOAG/3 & 0.266  & 0.237  & 0.252  & 0.262  & \underline{0.216} & 0.288  & 0.245  & \underline{0.216} & Chlor/5 & 0.245  & 0.241  & 0.340  & 0.161  & 0.217  & 0.131  & \underline{0.095} & \textbf{0.058} \\
DPOC/3 & \underline{0.210} & 0.225  & 0.243  & 0.234  & \underline{0.210} & 0.264  & 0.217  & \textbf{0.181} & EQ/5  & 0.273  & 0.252  & 0.252  & 0.330  & \textbf{0.180} & 0.288  & \underline{0.223} & 0.245  \\
DPTW/3 & 0.345  & \underline{0.281} & 0.324  & 0.336  & 0.288  & 0.302  & \textbf{0.268} & \underline{0.281} & FordA/5 & 0.052  & 0.044  & 0.050  & 0.045  & 0.075  & \underline{0.042} & 0.110  & \textbf{0.028} \\
FaceAll/3 & 0.193  & 0.118  & 0.158  & 0.155  & \textbf{0.072} & 0.182  & 0.098  & \underline{0.090} & FordB/5 & 0.180  & 0.163  & 0.177  & 0.162  & 0.225  & \underline{0.156} & 0.222  & \textbf{0.144} \\
FaceUCR/3 & 0.040  & 0.035  & \underline{0.033} & \underline{0.033} & 0.071  & 0.038  & 0.087  & 0.037  & ItaPD/5 & 0.037  & \underline{0.030} & 0.035  & 0.053  & \underline{0.030} & 0.041  & \textbf{0.023} & \underline{0.030} \\
FWords/3 & \underline{0.163} & 0.167  & \textbf{0.152} & 0.184  & 0.198  & 0.169  & 0.281  & 0.167  & Lgt2/5 & 0.246  & 0.213  & 0.164  & 0.193  & \textbf{0.049} & 0.115  & 0.145  & \underline{0.066} \\
HandO/3 & 0.065  & 0.059  & 0.062  & 0.071  & 0.051  & \underline{0.046} & 0.100  & \textbf{0.024} & Lgt7/5 & 0.205  & 0.192  & 0.233  & 0.207  & \underline{0.123} & 0.219  & \textbf{0.091} & 0.178  \\
Herring/3 & \underline{0.312} & 0.391  & 0.359  & 0.392  & \textbf{0.297} & 0.484  & 0.453  & 0.329  & Phonm/5 & 0.706  & \underline{0.633} & 0.639  & 0.695  & 0.715  & 0.698  & 0.800  & \textbf{0.604} \\
MedIm/3 & 0.209  & 0.193  & 0.204  & 0.231  & 0.207  & 0.212  & \textbf{0.164} & \underline{0.190} & SAR1/5 & 0.113  & 0.085  & 0.168  & \textbf{0.020} & 0.065  & \underline{0.022} & 0.042  & 0.065  \\
MPOAG/3 & 0.448  & 0.422  & 0.429  & 0.464  & \textbf{0.331} & 0.539  & 0.461  & \underline{0.351} & SAR2/5 & 0.079  & 0.077  & 0.104  & \textbf{0.046} & 0.075  & 0.072  & 0.064  & \underline{0.049} \\
MPOC/3 & \underline{0.151} & \underline{0.151} & 0.175  & 0.186  & \textbf{0.144} & 0.189  & 0.192  & 0.172  & SLCrv/5 & 0.018  & \underline{0.017} & \underline{0.017} & 0.025  & 0.023  & 0.023  & 0.018  & \underline{0.017} \\
\cline{10-18}MPTW/3 & 0.474  & 0.422  & 0.435  & 0.481  & \underline{0.396} & 0.519  & 0.429  & \textbf{0.364} & TP/6  & \underline{0.003}  & \textbf{0} & \textbf{0} & \textbf{0} & \textbf{0} & \textbf{0} & \textbf{0} & \textbf{0} \\
OSULf/3 & 0.041  & 0.033  & \textbf{0.004} & 0.060  & 0.140  & 0.037  & \underline{0.018} & 0.029  & CBF/6 & 0.083  & \textbf{0} & \underline{0.002}  & \textbf{0} & \underline{0.002}  & \textbf{0} & \textbf{0} & \textbf{0} \\
POC/3 & \textbf{0.160} & 0.162  & 0.182  & 0.170  & \underline{0.161} & 0.174  & 0.163  & \underline{0.161} & Mallt/6 & 0.054  & 0.028  & 0.029  & 0.036  & \underline{0.026} & 0.032  & 0.044  & \textbf{0.018} \\
\cline{10-18}PPOAG/3 & 0.166  & 0.146  & 0.141  & 0.156  & \underline{0.127} & 0.161  & \textbf{0.120} & 0.132  & Beef/7 & 0.167  & 0.167  & 0.233  & 0.193  & 0.133  & \underline{0.100} & \textbf{0.070} & \underline{0.100} \\
PPOC/3 & 0.096  & 0.100  & 0.117  & 0.092  & 0.086  & \underline{0.079} & 0.107  & \textbf{0.053} & Ham/7 & 0.286  & 0.276  & 0.295  & 0.296  & \textbf{0.181} & 0.286  & 0.257  & \underline{0.219} \\
PPTW/3 & 0.200  & 0.171  & 0.176  & 0.227  & 0.171  & 0.215  & \underline{0.166} & \textbf{0.161} & Olive/7 & \underline{0.067} & 0.133  & 0.100  & 0.213  & \underline{0.067} & 0.133  & \underline{0.067} & 0.133  \\
ShAll/3 & 0.077  & 0.082  & \underline{0.075} & 0.080  & 0.122  & 0.085  & 0.133  & \textbf{0.065} & Straw/7 & \underline{0.016} & 0.022  & 0.030  & 0.018  & 0.022  & \underline{0.016} & 0.049  & \underline{0.016} \\
Yoga/3 & 0.090  & \textbf{0.071} & 0.146  & \underline{0.089} & 0.133  & 0.100  & 0.112  & 0.090  & Wine/7 & 0.148  & \underline{0.074} & 0.130  & 0.256  & \underline{0.074} & 0.204  & 0.093  & \underline{0.074} \\
\bottomrule
\end{tabular}%
}\vspace{-4mm}\label{tab:clf_main}%
\end{table*}%

\section{Experimental Results}\label{sec:predexp}

\subsection{Task I: Time Series Classification}\label{subsec:Classification}

\subsubsection{Datasets} We evaluate our model on the UCR TSC archive 2018~\cite{dau2019ucr}, which is a commonly used large-scale univariate time series dataset covering the areas of image contour classification, motion classification, ECG classification, sensor data classification, and others. {The UCR archive is designed to provide a standard and comprehensive archive for univariate TSC, which has split the data into training and test set.} Due to space limitations, we select 64 representative datasets from the UCR archive in our experiments.
The dataset selection follows two principles. First, the selected representative datasets are intended to provide a more challenging and discriminative benchmark for evaluating different methods. Many UCR datasets have been extensively studied, and their performance is nearly saturated. That is, many algorithms achieve close-to-zero error rates on these datasets. We therefore excluded these nearly saturated datasets from the representative subset. Second, the selected subset covers all major UCR data categories used in our experiments, including Device, ECG, Image, Motion, Sensor, Simulated, and Spectro datasets. {\highlight In addition, we also provide the experimental results on all 128 UCR datasets in  Sec. 3 of SM.}

\subsubsection{Evaluation Metric and Baselines}We choose the error rate to evaluate the classification performance. We utilize seven baselines for comparison, including a feature-based approach, two ensemble methods, three deep learning models, and a wavelet-based deep model.

\begin{itemize}[leftmargin=*]
    \item {\textbf{miniRocket}~\cite{dempster2021minirocket}}: An advanced model that utilizes convolution kernels to extract features from time series to improve classification outcomes. Using a ridge regression classifier, miniRocket demonstrates superior performance compared to traditional techniques in TSC. We chose it as the exemplar baseline for feature-based approaches.
    \item {\textbf{Hivecote2}~\cite{middlehurst2021hive}}: A heterogeneous meta ensemble model that achieves state-of-the-art performance in TSC. It is an improved version of Hivecote that forms its ensemble from classifiers of multiple domains.
    \item {\textbf{TS-Chief}~\cite{TS-CHIEF}}: TS-Chief, short for Time Series Combination of Heterogeneous and Ensemble Embedding Forests, leverages the scalability of tree classifiers and decades of research into accurate and specialized TSC techniques.
    \item {\textbf{OS-CNN}~\cite{tang2020omni}}: It proposes an omni-scale block (OS-block) for 1D CNN, which uses many kernels of different sizes for multi-scale feature extraction. It is a representative of CNN-based deep learning models.
    \item {\textbf{TimeURL}~\cite{liu2024timesurl}}: A self-supervised model for time series representation learning that employs contrastive learning to capture both segment-level and instance-level information. It is the latest representative of deep learning models utilizing self-supervised techniques.
    \item {\textbf{MILLET}~\cite{early2023inherently}}: It is a recent deep learning model that employs multiple instance learning techniques to improve performance and provide local explanations for TSC.
    \item {\textbf{mWDN}~\cite{wang2018multilevel}}: This method focuses on the frequency-based decomposition. It uses a multi-level Wavelet Decomposition Network, which implements an approximate wavelet decomposition using a fully convolutional layer.
\end{itemize}



\subsubsection{Implementation Details}

Our \cls model uses the Adam optimizer for training. We configure the batch size to 16. The task balancing coefficients $\gamma_1, \gamma_2$ are adjusted using a dynamic weight-averaging method~\cite{liu2019end}, beginning with values of 1.0 each. The maximum number of training epochs is 200; during epochs 1 to 100, the learning rate is maintained at 0.001, decreases to 0.0001 between epochs 100 to 150, and further reduces to 0.00001 from epochs 150 to 200.
The number of decomposition levels is searched from 1 to 5, while the hidden dimension is searched in the set $\{16, 24, 32, 48\}$. These hyperparameters are set based on the optimal performance on the validation dataset, with details in Sec. 1.4 of SM. {\highlight Our code is available at \url{https://github.com/Beihang-BIGSCity/mwcn_ts}.}


\begin{table}
\centering
\caption{TSC performance summary. Avg. Err. denotes the average classification error rate. Count: winning count. Rank(a)/Rank(g): the average ranking in terms of arithmetic and geometry.}\vspace{-2mm}\label{tab:clf_sum}
\scriptsize 
\setlength{\tabcolsep}{1pt}
\begin{tabular}{c|cccccccc}
\toprule
Metric & mRocket & Hivecote2 & TS-Chief & OS-CNN & TimeURL & {MILLET} & mWDN  & \cls \\
\midrule
Avg. Err. & 0.187 & \underline{0.171} & 0.186 & 0.192 & 0.173 & 0.186 & 0.184 & \textbf{0.146} \\
Count & 3     & 7     & 6     & 7     & 12    & 5     & 11    & \textbf{34} \\
Rank (a) & 5.11  & 3.53  & 5.00  & 5.34  & 4.17  & 5.00  & 4.63  & \textbf{1.88 } \\
Rank (g) & 4.58  & 3.10  & 4.32  & 4.61  & 3.32  & 4.32  & 3.71  & \textbf{1.59 } \\
\bottomrule
\end{tabular}\vspace{-3mm}
\end{table}

\subsubsection{Results and Analysis}


\tab~\ref{tab:clf_main} shows the detailed experimental results of 64 UCR datasets, with a summary in \tab~\ref{tab:clf_sum}. Each experiment was run five times with different random seeds, and the average performance is reported. From the results, we can have the following key observations.

First, it is clear that among all the competitors, TFBC achieves the best performance in terms of both the largest number of wins (the best in 34 out of 64 datasets) and the highest average rank with regard to both arithmetic (1.88) and geometry (1.59). In addition, TFBC reduces the average error rate by 19.97\% over all baselines on average (See Sec.~4 of SM for the calculation details of average performance improvement).
The rank index indicates that even in the cases where our model is not the best, its performance is still very competitive. \cls's superior performance demonstrates the effectiveness of our model and underscores the importance of simultaneously capturing frequency and pattern information. Our model's superiority is further verified by the Nemenyi test at level 5\% in Sec. 1.3 of SM.

Second, our \cls beats other competitive models across all categories of datasets, highlighting its robustness and adaptability to different data types. Compared with the second-best model, \cls exhibits a greater improvement on datasets belonging to IMAGE, MOTION, and SENSOR types (refer to the illustration in Sec. 1.2 of SM). This can be attributed to the complex patterns of these datasets, where the performance benefits of multi-wavelet decomposition are more pronounced.

Third, mWDN, as a competitive baseline, can be seen as a degradation of \cls that removes the prefiltering module that extracts distinct patterns. The comparison between mWDN and \cls highlights that the pattern-based decomposition is essential to the success of \cls in TSC.

Finally, deep learning methods, including TimeURL, mWDN, and \cls, perform better overall, indicating that deep learning's representation learning ability is suitable for extracting features from large-scale time series data. Besides, TimeURL, which adopts self-supervised learning technology, achieves better results by refining the representation through contrastive learning. This is why we also incorporate contrastive learning into model training.

{\highlight The results on the full set of 128 UCR datasets lead to consistent observations (See Sec. 3 of SM).}. 

\begin{table}
\centering
\caption{\highlight Ablation study on the 64 UCR datasets. Avg. Err. denotes the average classification error rate. Count: winning count. Rank(a)/Rank(g): the average ranking in terms of arithmetic and geometry.}\vspace{-2mm}\label{tab:ablation_ucr_64}
\scriptsize \highlight  
\setlength{\tabcolsep}{4.2pt}
\begin{tabular}{l|cccccc}
\toprule
Metric & r/fp & r/wd & w/o m-WCN & w/o OR & w/o FCL & TFBC \\
\midrule
Avg. Err. & 0.162 & 0.170 & 0.294 & 0.156 & \underline{0.154} & \textbf{0.146} \\
Count & 9 & 15 & 5 & 10 & \underline{14} & \textbf{26} \\
Rank(a) & 3.187 & 3.516 & 5.125 & 2.781 & \underline{2.828} & \textbf{2.281} \\
Rank(g) & 2.790 & 2.943 & 4.641 & 2.487 & \underline{2.471} & \textbf{1.921} \\
\bottomrule
\end{tabular}\vspace{-3mm}
\end{table}

\subsubsection{Ablation of Important Modules}\label{sec:tsc_ab}

{\highlight We conduct an ablation study on the 64 representative UCR datasets to evaluate the contribution of each proposed component.} We compare TFBC with five variants:
(1) \textit{r/fp}, which replaces the learnable pre-filtering network with fixed GHM pre-filtering parameters;
(2) \textit{r/wd}, which replaces the learnable multi-wavelet decomposition with standard wavelet decomposition;
(3) \textit{w/o m-WCN}, which removes the m-WCN module;
(4) \textit{w/o OR}, which removes the orthogonality regularization in Eq.~\eqref{eq:orthogonal regularization};
and (5) \textit{w/o FCL}, which removes the frequency contrastive learning loss in Eq.~\eqref{eq:contrastive learning}.
The results are summarized in Table~\ref{tab:ablation_ucr_64}.

{\highlight The results show that each component contributes to the final performance. {TFBC achieves the lowest average error rate of 0.146, the largest winning count of 26, and the best arithmetic and geometric average ranks of 2.281 and 1.921, respectively. Removing the whole m-WCN module leads to the largest performance drop. The average error rate increases from 0.146 to 0.294, and the arithmetic average rank worsens from 2.281 to 5.125.} This confirms that the neuralized multi-wavelet decomposition module is the core component of TFBC.

The comparison with \textit{r/fp} and \textit{r/wd} further verifies the benefit of learnable multi-wavelet decomposition. When the learnable pre-filtering network is replaced with fixed parameters, the average error rate increases to 0.162. When multi-wavelet decomposition is replaced by standard wavelet decomposition, the average error rate increases to 0.170. These results show that both learnable pattern decomposition and multi-wavelet frequency decomposition are important. They allow TFBC to adapt the decomposition process to different datasets, rather than using fixed decomposition only as feature engineering.

The results of \textit{w/o OR} and \textit{w/o FCL} also demonstrate the usefulness of the two training regularizers. Removing orthogonality regularization increases the average error rate to 0.156 and reduces the winning count to 10. Removing frequency contrastive learning increases the average error rate to 0.154 and reduces the winning count to 14. These results show that orthogonality regularization helps reduce redundancy among decomposed components, while frequency contrastive learning improves the discriminative ability of frequency representations. Overall, the ablation results confirm that m-WCN, learnable decomposition, orthogonality regularization, and frequency contrastive learning all make positive contributions to TFBC.

The ablation results on the full 128 UCR datasets are provided in Sec. 3.2 of SM. The conclusions are consistent with those obtained on the 64 representative datasets.}


\subsection{Task II: Time Series Forecasting}\label{subsec:Forecasting}

\subsubsection{Datasets}

In the experiments, we compare our model with baseline models over {seven} real-world TSF datasets. All of the datasets are publicly available:
\begin{itemize}[leftmargin=*]
    \item ETT (h1, h2, m1, m2): the Electricity Transformer Temperature (ETT) datasets~\cite{zhou2021informer} contain 2 years of electricity transformer temperature data from a county in China. They include four subsets: ETTh1 and ETTh2 at the 1-hour level, and ETTm1 and ETTm2 at the 15-minute level.

%
%
%
    \item Electricity: the electricity dataset from the UCI Machine Learning Repository\footnote{\url{https://archive.ics.uci.edu/ml/datasets}} contains hourly electricity consumption for 370 clients from 2012 to 2014.

    \item Traffic: the traffic dataset from the California Department of Transportation\footnote{\url{http://pems.dot.ca.gov}} contains road occupancy rates measured by 862 sensors in the San Francisco Bay area freeways during 2015 and 2016.

    \item {\highlight Weather: the weather dataset from NOAA/NCEI Local Climatological Data\footnote{\url{https://www.ncei.noaa.gov/data/local-climatological-data/}} contains local climatological observations from nearly 1,600 U.S. locations over four years from 2010 to 2013. The data are collected at 1-hour intervals. Each data point contains the target value ``wet bulb'' and 11 climate features.}

\end{itemize}
In the experiments, we follow common task settings in TSF and use our model and baselines to forecast the future 96, 192, 336, and 720 steps. The input sequence length for each forecasting task is searched over the set $\{96,192,336,720\}$. {The original series is divided into training, validation, and test sets in a ratio of 6:2:2 for ETT Datasets and in a ratio of 7:1:2 for other datasets.} Subsequently, a sliding window approach is employed to produce data samples for each dataset.

\subsubsection{Evaluation Metric and Baselines}
In the experiments, we use two common metrics to evaluate prediction accuracy: Mean Squared Error (MSE) and Mean Absolute Error (MAE). We compared our model with {seven} competitive baselines.
\begin{itemize}[leftmargin=*]
    \item {\textbf{Autoformer}~\cite{wu2021autoformer}}: The approach introduces a decomposition mechanism grounded in auto-correlation. It adheres to the Transformer encoder-decoder framework but incorporates a decomposition module to capture the complex temporal dynamics of the hidden states.
    \item {\textbf{Fedformer}~\cite{zhou2022fedformer}}: This model is a frequency-enhanced decomposed transformer utilizing the seasonal-trend decomposition approach. The decomposition mechanism encapsulates the overall pattern of time series data, while Transformers extract detailed structural aspects.
    \item {\textbf{Dlinear}~\cite{zeng2023transformers}}: The model is a combination of a decomposition scheme used in Autoformer and Fedformer with linear layers. It first decomposes the input series into a trend component and a remainder (seasonal) component, and then two linear layers are applied to each component for the final prediction.
    \item {\textbf{Basisformer}~\cite{ni2024basisformer}}: This model utilizes cross-attention to calculate the similarity coefficients between the time series and learnable bases in the historical view, and then selects the bases in the future view based on the similarity coefficients for accurate prediction.
    \item {\textbf{Pathformer}~\cite{chen2024pathformer}}: It models the multi-scale characteristics of time series with adaptive pathways integrating temporal resolutions and temporal distance information.
    \item {\textbf{PatchTST}~\cite{nie2022time}}: This method segments time series into subseries-level patches, which are served as input tokens to the Transformer.
    \item {\highlight \textbf{TimeLLM}~\cite{timellm}}: {\highlight This is a recent large language model based framework for time series forecasting. It reformulates time series forecasting as a sequence modeling task and leverages the representation and reasoning ability of pre-trained language models. 
        We include TimeLLM as a recent strong baseline to evaluate whether FTB remains competitive against LLM-based forecasting methods.}
\end{itemize}


\begin{table}[t]
\centering
\caption{\highlight Performance comparison of TSF with different prediction lengths. The best result is bold, while the second-best result is underlined. Auto.=Autoformer, Fed.=Fedformer, Basis.=Basisformer, and Path.=PathFormer.}
\label{tab:forecasting}
\vspace{-2mm}
{\highlight%
\setlength{\tabcolsep}{0.5mm}
\renewcommand{\arraystretch}{1.1}
\tiny
\begin{tabular}{c|c|c|cccccccc}
\toprule
Dataset & Horizon & Metric & Auto. & Fed. & DLinear & Basis. & Path. & TimeLLM & PatchTST & FTB \\
\midrule
\multirow{8}{*}{\rotatebox[origin=c]{90}{ETTh1}} & \multirow{2}{*}{96} & MSE & 0.071 & 0.079 & 0.056 & \underline{0.055} & 0.057 & 0.058 & 0.057 & \textbf{0.052} \\
 &  & MAE & 0.206 & 0.215 & 0.180 & \underline{0.178} & 0.180 & 0.183 & 0.179 & \textbf{0.165} \\
 & \multirow{2}{*}{192} & MSE & 0.114 & 0.104 & \underline{0.071} & 0.072 & 0.075 & 0.072 & 0.076 & \textbf{0.064} \\
 &  & MAE & 0.262 & 0.245 & 0.204 & 0.204 & 0.208 & \underline{0.202} & 0.209 & \textbf{0.180} \\
 & \multirow{2}{*}{336} & MSE & 0.107 & 0.119 & 0.098 & 0.086 & \underline{0.076} & 0.082 & 0.093 & \textbf{0.063} \\
 &  & MAE & 0.258 & 0.270 & 0.244 & 0.227 & \underline{0.216} & 0.231 & 0.240 & \textbf{0.201} \\
 & \multirow{2}{*}{720} & MSE & 0.126 & 0.142 & 0.189 & \underline{0.080} & 0.090 & 0.093 & 0.097 & \textbf{0.077} \\
 &  & MAE & 0.283 & 0.299 & 0.359 & \underline{0.220} & 0.238 & 0.243 & 0.245 & \textbf{0.213} \\
\midrule
\multirow{8}{*}{\rotatebox[origin=c]{90}{ETTh2}} & \multirow{2}{*}{96} & MSE & 0.150 & 0.132 & 0.132 & 0.133 & 0.137 & 0.132 & \textbf{0.129} & \underline{0.130} \\
 &  & MAE & 0.303 & 0.287 & \textbf{0.279} & 0.286 & 0.291 & 0.286 & \underline{0.282} & 0.284 \\
 & \multirow{2}{*}{192} & MSE & 0.195 & 0.171 & 0.175 & 0.183 & 0.371 & 0.177 & \textbf{0.168} & \underline{0.170} \\
 &  & MAE & 0.343 & \underline{0.331} & 0.334 & 0.336 & 0.390 & 0.336 & \textbf{0.328} & \underline{0.331} \\
 & \multirow{2}{*}{336} & MSE & 0.234 & 0.193 & 0.211 & 0.211 & 0.331 & 0.195 & \underline{0.185} & \textbf{0.183} \\
 &  & MAE & 0.387 & 0.366 & 0.369 & 0.367 & 0.373 & 0.374 & \underline{0.351} & \textbf{0.346} \\
 & \multirow{2}{*}{720} & MSE & 0.272 & 0.233 & 0.295 & 0.238 & 0.417 & 0.237 & \underline{0.224} & \textbf{0.219} \\
 &  & MAE & 0.418 & 0.387 & 0.442 & 0.393 & 0.434 & 0.397 & \underline{0.383} & \textbf{0.371} \\
\midrule
\multirow{8}{*}{\rotatebox[origin=c]{90}{ETTm1}} & \multirow{2}{*}{96} & MSE & 0.051 & 0.029 & 0.027 & 0.029 & 0.029 & 0.029 & \underline{0.026} & \textbf{0.022} \\
 &  & MAE & 0.176 & 0.127 & 0.123 & 0.127 & 0.125 & 0.129 & \underline{0.121} & \textbf{0.116} \\
 & \multirow{2}{*}{192} & MSE & 0.076 & 0.042 & 0.043 & 0.044 & 0.040 & 0.044 & \underline{0.039} & \textbf{0.031} \\
 &  & MAE & 0.221 & 0.159 & 0.154 & 0.160 & 0.153 & 0.162 & \underline{0.150} & \textbf{0.139} \\
 & \multirow{2}{*}{336} & MSE & 0.081 & 0.059 & 0.060 & 0.059 & 0.057 & 0.062 & \underline{0.053} & \textbf{0.042} \\
 &  & MAE & 0.226 & 0.185 & 0.180 & 0.186 & 0.182 & 0.188 & \underline{0.173} & \textbf{0.162} \\
 & \multirow{2}{*}{720} & MSE & 0.106 & 0.079 & 0.081 & 0.083 & 0.082 & 0.085 & \underline{0.074} & \textbf{0.059} \\
 &  & MAE & 0.258 & 0.214 & 0.211 & 0.221 & 0.220 & 0.227 & \underline{0.207} & \textbf{0.189} \\
\midrule
\multirow{8}{*}{\rotatebox[origin=c]{90}{ETTm2}} & \multirow{2}{*}{96} & MSE & 0.086 & 0.066 & 0.070 & 0.071 & \underline{0.064} & 0.073 & 0.070 & \textbf{0.062} \\
 &  & MAE & 0.223 & 0.193 & 0.191 & 0.191 & \underline{0.181} & 0.196 & 0.191 & \textbf{0.177} \\
 & \multirow{2}{*}{192} & MSE & 0.148 & 0.113 & 0.104 & 0.104 & \textbf{0.100} & 0.106 & 0.104 & \underline{0.102} \\
 &  & MAE & 0.295 & 0.262 & 0.238 & 0.235 & \textbf{0.232} & 0.241 & 0.238 & \underline{0.234} \\
 & \multirow{2}{*}{336} & MSE & 0.155 & 0.158 & 0.135 & 0.130 & \underline{0.129} & 0.132 & 0.135 & \textbf{0.126} \\
 &  & MAE & 0.309 & 0.305 & 0.278 & 0.275 & \underline{0.270} & 0.273 & 0.278 & \textbf{0.260} \\
 & \multirow{2}{*}{720} & MSE & \underline{0.177} & 0.199 & 0.188 & 0.182 & 0.180 & 0.183 & 0.188 & \textbf{0.158} \\
 &  & MAE & 0.329 & 0.348 & 0.332 & 0.332 & \underline{0.329} & 0.330 & 0.332 & \textbf{0.315} \\
\midrule
\multirow{8}{*}{\rotatebox[origin=c]{90}{Electricity}} & \multirow{2}{*}{96} & MSE & 0.381 & 0.253 & 0.374 & 0.333 & 0.373 & 0.221 & \underline{0.217} & \textbf{0.203} \\
 &  & MAE & 0.458 & 0.370 & 0.439 & 0.408 & 0.520 & 0.335 & \underline{0.326} & \textbf{0.322} \\
 & \multirow{2}{*}{192} & MSE & 0.469 & 0.282 & 0.351 & 0.371 & 0.306 & 0.269 & \underline{0.256} & \textbf{0.247} \\
 &  & MAE & 0.516 & 0.386 & 0.422 & 0.427 & 0.389 & 0.384 & \underline{0.354} & \textbf{0.351} \\
 & \multirow{2}{*}{336} & MSE & 0.591 & 0.346 & 0.379 & 0.413 & 0.369 & 0.322 & \underline{0.296} & \textbf{0.285} \\
 &  & MAE & 0.570 & 0.431 & 0.442 & 0.455 & 0.427 & 0.417 & \underline{0.387} & \textbf{0.378} \\
 & \multirow{2}{*}{720} & MSE & 0.658 & 0.422 & 0.417 & 0.471 & 0.383 & 0.356 & \underline{0.339} & \textbf{0.323} \\
 &  & MAE & 0.606 & 0.484 & 0.478 & 0.498 & 0.443 & 0.448 & \underline{0.431} & \textbf{0.421} \\
\midrule
\multirow{8}{*}{\rotatebox[origin=c]{90}{Traffic}} & \multirow{2}{*}{96} & MSE & 0.244 & 0.207 & 0.303 & 0.186 & \underline{0.144} & 0.164 & 0.176 & \textbf{0.102} \\
 &  & MAE & 0.352 & 0.312 & 0.396 & 0.280 & \underline{0.211} & 0.246 & 0.253 & \textbf{0.172} \\
 & \multirow{2}{*}{192} & MSE & 0.280 & 0.205 & 0.247 & 0.175 & \underline{0.139} & 0.171 & 0.162 & \textbf{0.122} \\
 &  & MAE & 0.385 & 0.312 & 0.335 & 0.267 & \underline{0.206} & 0.250 & 0.243 & \textbf{0.189} \\
 & \multirow{2}{*}{336} & MSE & 0.271 & 0.219 & 0.241 & 0.179 & \underline{0.141} & 0.153 & 0.164 & \textbf{0.125} \\
 &  & MAE & 0.374 & 0.323 & 0.329 & 0.273 & \underline{0.212} & 0.222 & 0.248 & \textbf{0.206} \\
 & \multirow{2}{*}{720} & MSE & 0.258 & 0.244 & 0.289 & 0.194 & \underline{0.181} & 0.185 & 0.189 & \textbf{0.161} \\
 &  & MAE & 0.365 & 0.344 & 0.370 & 0.284 & \underline{0.248} & 0.251 & 0.267 & \textbf{0.232} \\
\midrule
\multirow{8}{*}{\rotatebox[origin=c]{90}{Weather}} & \multirow{2}{*}{96} & MSE & 0.0176 & 0.0101 & 0.0056 & 0.0014 & \underline{0.0012} & 0.0016 & 0.0012 & \textbf{0.0011} \\
 &  & MAE & 0.1017 & 0.0808 & 0.0620 & 0.0277 & \underline{0.0241} & 0.0324 & 0.0257 & \textbf{0.0241} \\
 & \multirow{2}{*}{192} & MSE & 0.0111 & 0.0053 & 0.0061 & 0.0016 & 0.0014 & 0.0014 & \underline{0.0014} & \textbf{0.0011} \\
 &  & MAE & 0.0735 & 0.0592 & 0.0658 & 0.0300 & \underline{0.0261} & 0.0283 & 0.0271 & \textbf{0.0255} \\
 & \multirow{2}{*}{336} & MSE & 0.0258 & 0.0058 & 0.0064 & 0.0017 & \underline{0.0015} & 0.0016 & 0.0015 & \textbf{0.0015} \\
 &  & MAE & 0.1028 & 0.0611 & 0.0674 & 0.0314 & \underline{0.0277} & 0.0291 & 0.0289 & \textbf{0.0144} \\
 & \multirow{2}{*}{720} & MSE & 0.0139 & 0.0072 & 0.0067 & 0.0022 & \underline{0.0019} & 0.0021 & 0.0020 & \textbf{0.0019} \\
 &  & MAE & 0.0705 & 0.0674 & 0.0693 & 0.0352 & \underline{0.0320} & 0.0337 & 0.0335 & \textbf{0.0306} \\
\bottomrule
\end{tabular}
\label{tab:class_performance_TSF}%
}
\vspace{-3mm}
\end{table}

\subsubsection{Implementation Details}
We have implemented the \ftb model utilizing PyTorch, and it is trained with the Adam optimizer at a learning rate of 0.001 for a total of 70 epochs. The hidden embedding size is configured to 64, with a batch size of 32. The task balancing coefficient $\gamma_3$ is optimized through a dynamic weight-averaging method~\cite{liu2019end}, starting from an initial value of 1.0.
We search the number of decomposition levels from 1 to 5, and the hidden dimension in the set $\{16, 24, 32, 48\}$. Both hyperparameters are set based on the optimal performance on validation datasets (see Sec. 2.1 of SM). {\highlight Our code is available at \url{https://github.com/Beihang-BIGSCity/mwcn_ts}.}

\subsubsection{Results and Analysis}

\tab~\ref{tab:class_performance_TSF} presents the comparison results of all the methods, from which we make the following three observations. {\highlight Moreover, \tab~\ref{tab:forecasting_summary} gives the performance summary over seven forecasting datasets of the TSF experiments.

First, according to \tab~\ref{tab:forecasting_summary}, our model consistently outperforms all competing baselines across most tasks on the seven datasets, with an average error reduction of 24.45\% in MSE and 15.38\% in MAE (See Sec.~4 of SM for the calculation details of average performance improvement). In contrast, the second-best model varies across tasks. This shows that our model offers more stable and reliable results, highlighting its robustness and adaptability to various data.}

Second, compared with Fedformer, which mainly exploits frequency-domain information, and PathFormer and PatchTST, which focus more on temporal pattern modeling, FTB jointly models pattern and frequency information through neuralized multi-wavelet decomposition. The results show that FTB achieves the best overall average MSE and MAE across the seven forecasting datasets. This supports the effectiveness of jointly optimizing pattern and frequency decomposition for time series forecasting.


Third, models with explicit structural priors often show more interpretable behavior as the prediction horizon changes. In our results, FTB generally exhibits a smooth increase in prediction error when the horizon becomes longer, which is consistent with the increasing difficulty of long-term forecasting. This behavior suggests that the proposed multi-wavelet decomposition provides a useful inductive bias for capturing changes in data predictability. By contrast, several baselines show more dataset-dependent trends across horizons.


\begin{table}[t]
\scriptsize\highlight
\centering
\caption{\highlight TSF performance summary over seven forecasting datasets. Count denotes the number of wins including ties over all metric-horizon comparisons. Rank(a) and Rank(g) denote the arithmetic and geometric average ranks.}
\label{tab:forecasting_summary}
    \setlength{\tabcolsep}{0.7mm}
    \renewcommand{\arraystretch}{1.1}
\begin{tabular}{l|cccccccc}
\toprule
Metric & Auto. & Fed. & DLinear & Basis. & Path. & TimeLLM & PatchTST & FTB \\
\midrule
Avg. MSE & 0.192 & 0.147 & 0.163 & 0.146 & 0.153 & 0.129 & \underline{0.126} & \textbf{0.112} \\
Avg. MAE & 0.303 & 0.265 & 0.271 & 0.248 & 0.242 & 0.238 & \underline{0.233} & \textbf{0.216} \\
Count & 0 & 0 & 1 & 0 & 5 & 0 & \underline{4} & \textbf{50} \\
Rank(a) & 7.42 & 5.61 & 5.55 & 4.87 & 3.72 & 4.49 & \underline{3.17} & \textbf{1.17} \\
Rank(g) & 7.26 & 5.35 & 5.22 & 4.65 & 3.21 & 4.26 & \underline{2.86} & \textbf{1.12} \\
\bottomrule
\end{tabular}
\end{table}

\subsubsection{Ablation of Important Modules}

The ablation study is conducted to analyze how each of the proposed components in \ftb affects the final forecasting performance. We prepare five variants for comparison. The first four variants are the same as those of Sec.~\ref{sec:tsc_ab}. The last variant removes the pre-training phase, denoted as \textit{w/o pt}. We report the experimental results on all forecasting tasks regarding both metrics in \tab~\ref{tab:tsf_ab}. Results on all {seven} datasets show a similar phenomenon, so we only report the results on the Electricity dataset for simplicity.
\begin{table}[t]
  \centering
  \caption{Ablation study of \ftb on the Electricity dataset.}\vspace{-2mm}
  \resizebox{0.8\linewidth}{!}{
    \setlength{\tabcolsep}{0.9mm}
    \renewcommand{\arraystretch}{1.1}
    \begin{tabular}{l|cc|cc|cc|cc}
    \toprule
    Horizon & \multicolumn{2}{c|}{96} & \multicolumn{2}{c|}{192} & \multicolumn{2}{c|}{336} & \multicolumn{2}{c}{720} \\
    \midrule
    Metric & MSE   & MAE   & MSE   & MAE   & MSE   & MAE   & MSE   & MAE \\
    \midrule
    \ftb & \textbf{0.203} & \textbf{0.322} & \textbf{0.247} & \textbf{0.351} & \textbf{0.285} & \textbf{0.378} & \textbf{0.323} & \textbf{0.421} \\
    r/ fp & 0.215  & 0.352  & 0.266  & 0.386  & 0.288  & 0.359  & 0.342  & 0.432  \\
    r/ wd & 0.247  & 0.359  & 0.319  & 0.366  & 0.340  & 0.389  & 0.373  & 0.486  \\
    w/o mwcn & 0.256  & 0.389  & 0.307  & 0.395  & 0.332  & 0.425  & 0.388  & 0.502  \\
    w/o or & 0.238  & 0.333  & 0.258  & 0.362  & 0.288  & 0.381  & 0.359  & 0.453  \\
    w/o pt & 0.228  & 0.341  & 0.268  & 0.368  & 0.315  & 0.391  & 0.345  & 0.488  \\
    \bottomrule
    \end{tabular}}%
  \vspace{-4mm}\label{tab:tsf_ab}%
\end{table}%

We can observe from \tab~\ref{tab:tsf_ab} that all components contribute to the model’s overall performance.
Particularly, variants \textit{w/o mwcn} and \textit{r/ wd} show a great performance decrease, indicating that our proposed \model can effectively capture the characteristics of the input series data by using learnable multi-wavelets for pattern and frequency decomposition.
In addition, the pre-training phase plays a crucial role in maintaining the model's performance. This is due to the fact that the pre-training task, which involves recovering the \mwd decomposition outcomes, imposes constraints on the model's learning process and reduces the risk of overfitting.

{\highlight
\subsubsection{Efficiency Comparison}

We further conduct empirical efficiency comparisons on seven forecasting datasets, including ETTh1, ETTh2, ETTm1, ETTm2, Electricity, Weather, and Traffic. All experiments are conducted on the same hardware platform: {a single NVIDIA GeForce RTX 3090 GPU}. For all methods, the input length and prediction length are both set to 336. Table~\ref{tab:efficiency} reports the average training time, training memory, inference time, and inference memory over the seven datasets. The training time is measured as the one-epoch training time, while the inference time is measured on the test set. As shown in Table~\ref{tab:efficiency}, FTB requires 63.76 seconds per training epoch, which is comparable to Autoformer and PathFormer. Its inference time is 6.24 seconds, also close to PathFormer. In terms of memory usage, FTB requires 464.07 MB during training and 407.62 MB during inference, which remains within a practical range among Transformer-based forecasting models. Overall, these results show that FTB maintains practical computational efficiency while achieving strong forecasting performance.
}

\begin{table}[t]
\centering
\caption{\highlight Average computational cost on the forecasting experiments. TT, TM, IT, and IM denote training time, training memory, inference time, and inference memory, respectively.}
\label{tab:efficiency}
\scriptsize \highlight
\setlength{\tabcolsep}{6pt}
\begin{tabular}{lcccc}
\toprule
Method & TT (s/epoch) & TM (MB) & IT (s) & IM (MB) \\
\midrule
Autoformer & 66.27 & 3848.13 & 22.53 & 738.82 \\
Fedformer & 237.78 & 2612.12 & 15.32 & 677.35 \\
DLinear & 7.54 & 21.14 & 4.62 & 19.63 \\
Basisformer & 42.49 & 58.47 & 3.52 & 39.15 \\
PathFormer & 58.39 & 2227.41 & 6.82 & 367.09 \\
PatchTST & 8.91 & 153.49 & 3.00 & 68.23 \\
\midrule
FTB & 63.76 & 464.07 & 6.24 & 407.62 \\
\bottomrule
\end{tabular}
\end{table}

\subsubsection{Prediction Visualization}\label{sec:pred_vis}

\begin{figure}[t]
    \centering
    \includegraphics[width=0.9\columnwidth]{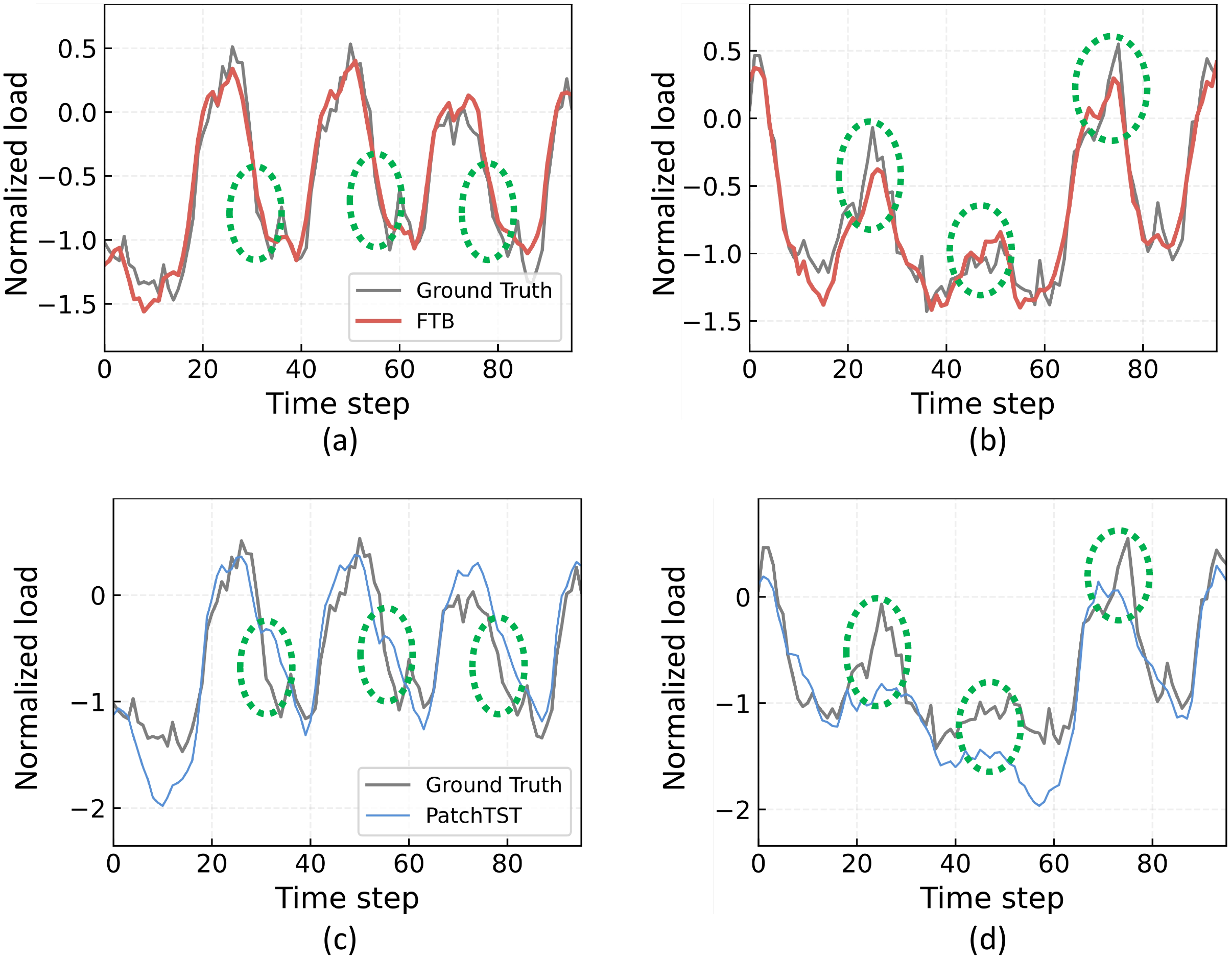}\vspace{-3mm}
    \caption{\highlight Visualization of forecasting results. Subfigures (a) and (b) show the prediction results of FTB, while subfigures (c) and (d) show the prediction results of PatchTST on the same samples.} 
    \label{fig:case}
    \vspace{-0.3cm}
\end{figure}

In this part, we further visualize the prediction results to explore the effectiveness of our model. As shown in \fig~\ref{fig:case}, we depict the prediction results of the 96-step forecasting task on the Electricity dataset. The left and right figures visualize periodic and non-periodic samples, respectively. In order to make the results more convincing, we also include the optimal baseline model, PatchTST, as a comparison.

From the visualization results, we have three main findings.
(1) From the left half of \fig~\ref{fig:case}, the periodic series mainly consists of two kinds of patterns, \ie rising and falling patterns. PatchTST accurately identifies the rising pattern, but it struggles with the falling pattern, as indicated by the dotted circle.
In contrast, our \ftb effectively captures both rising and falling patterns. This capability stems from our pattern decomposition module, which enables the features from different channels to discern unique patterns (refer to Sec. 2.2 of SM for decomposed component visualization), thereby enhancing pattern learning quality.
(2) In the more intricate non-periodic sample depicted in the right-hand section of \fig~\ref{fig:case}, PatchTST learned the trend shifts but significantly struggles to detect high-frequency details, as shown in dotted circles.
However, our \ftb, which utilizes frequency decomposition, proficiently captures both high- and low-frequency details, leading to highly accurate predictions. This highlights the efficacy of our model in utilizing frequency data to improve the analysis of complex time series.
(3) Integrating the findings from both subfigures reveals that our model delivers superior predictive performance on time series signals regardless of whether their distributions are periodic or aperiodic. This highlights the robustness of our model and its adaptability in diverse situations.

\section{Conclusion and Future Work}\label{sec:conclude}


In this paper, we proposed the Multi-Wavelet Decomposition Convolution Network (\model), a novel framework that integrates frequency and pattern decomposition within a trainable, end-to-end architecture for time series analysis. By incorporating neuralized MWD and orthogonality regularization, \model achieves enhanced adaptability while preserving interpretability. Moreover, it can be seamlessly integrated into deep learning pipelines. We developed two \model-based models, \cls and \ftb, specifically designed for classification and forecasting tasks, respectively.
Extensive experiments on diverse real-world datasets demonstrated our models' superiority to state-of-the-art baselines across various settings, underscoring our models' robustness and versatility in handling time series data.


Due to the characteristics of \mwd, our current framework is naturally suited for univariate time series. Extending it to multivariate time series requires handling each variable independently, which can be cumbersome. As part of future work, we aim to extend \model to support multivariate time series more naturally, thereby broadening its applicability and impact within the time series community.

\section*{Acknowledgments}
Prof. Wang's work is supported by the National Natural Science Foundation of China (No. 72242101, 72625015), and the Science and Technology Development Fund Macau SAR (0052/2023/RIA1). Dr. Junjie Wu's work was partially supported by the National Natural Science Foundation of China (72595861), the Outstanding Young Scientist Program of Beijing Universities (JWZQ20240201002), and the Shenzhen Science and Technology Program (CJGJZD20230724093201004).

\bibliographystyle{IEEEtran}
\bibliography{IEEEabrv,ref-sim}

 \vspace{-1.3cm}
\begin{IEEEbiography}[{\includegraphics[width=1in, height=1.25in,clip,keepaspectratio]{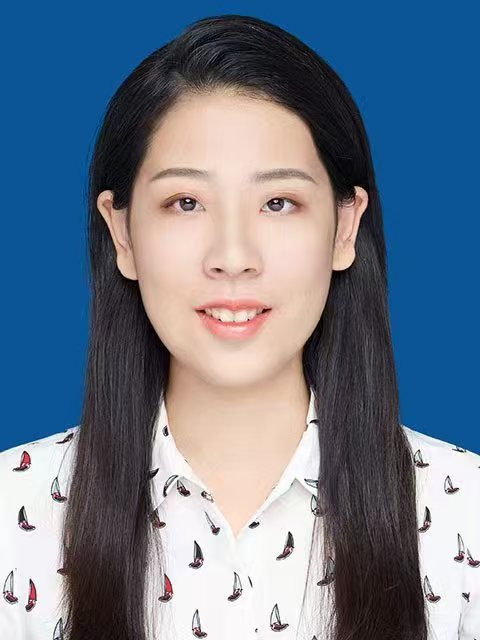}}]{Xiaohan Jiang}
is a PhD student at the School of Computer Science and Engineering, Beihang University, China. Her research interests include natural language processing, time series analysis, and interpretable machine learning.
\end{IEEEbiography}
 \vspace{-1.3cm}

\begin{IEEEbiography}[{\includegraphics[width=1in, height=1.25in,clip,keepaspectratio]{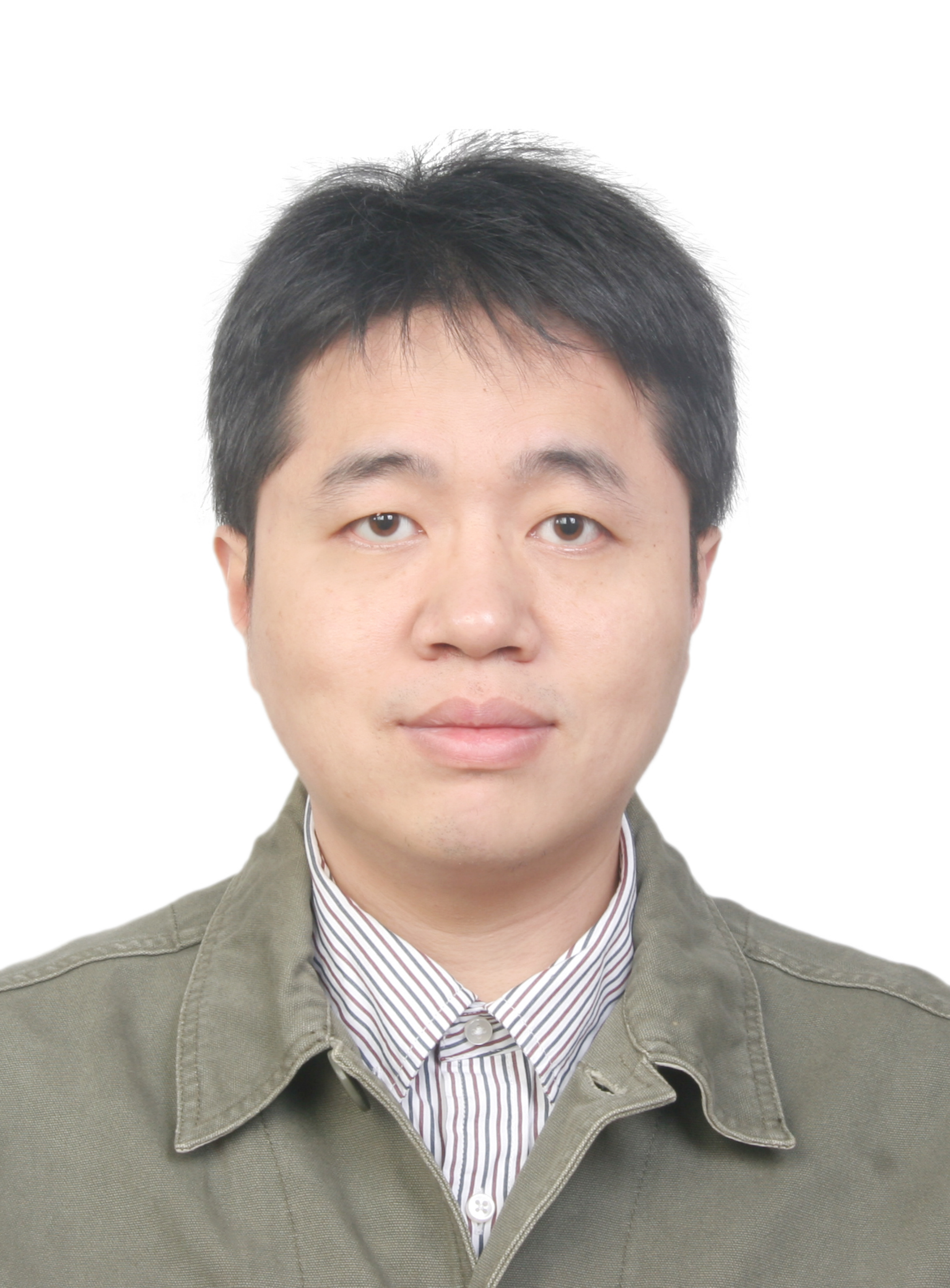}}]{Jingyuan Wang}
received his Ph.D. degree from the Department of Computer Science and Technology at Tsinghua University. He is currently a Professor at the School of Computer Science and Engineering and the School of Economics and Management, Beihang University. He is also the head of the Beihang Interest Group on SmartCity (BIGSCity). His research interests include data mining and machine learning, with a particular focus on smart cities and spatiotemporal data analytics. He has received several prestigious academic honors and awards, including the NSFC Distinguished Young Scholar, the NSFC for Excellent Young Scholar, the Beijing Young Scholar Award, and the First Prize of the Ministry of Education's Technological Invention Award.
\end{IEEEbiography}
 \vspace{-1.3cm}

\begin{IEEEbiography}[{\includegraphics[width=1in, height=1.25in,clip,keepaspectratio]{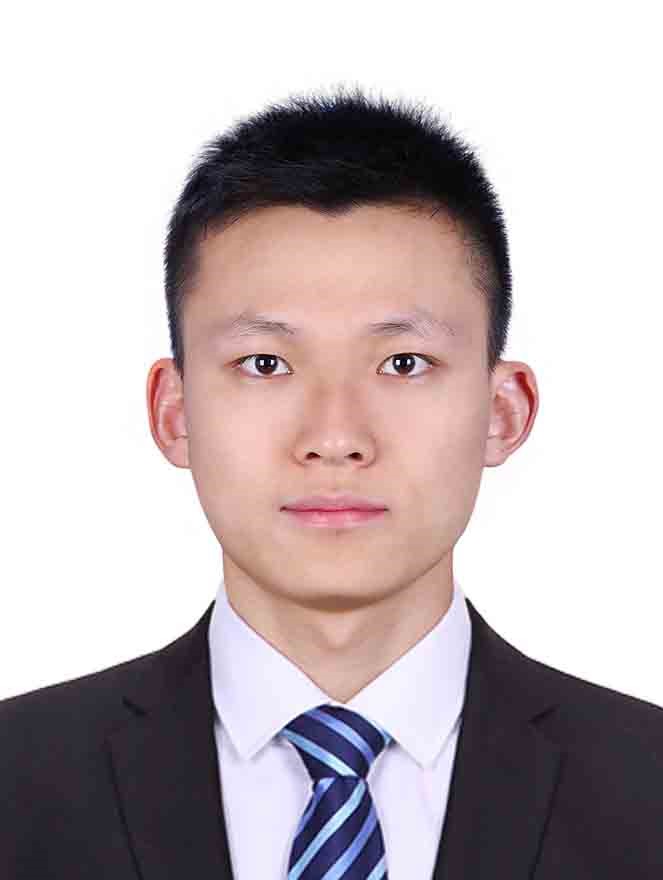}}]{Jiahao Ji}
is a Ph.D. candidate at the School of Computer Science and Engineering, Beihang University. He received his B.S from Beihang University in 2019.
His research interests include spatio-temporal data mining, interpretable machine learning, and urban computing.
\end{IEEEbiography}
 \vspace{-1.3cm}

\begin{IEEEbiography}[{\includegraphics[width=1in, height=1.25in,clip,keepaspectratio]{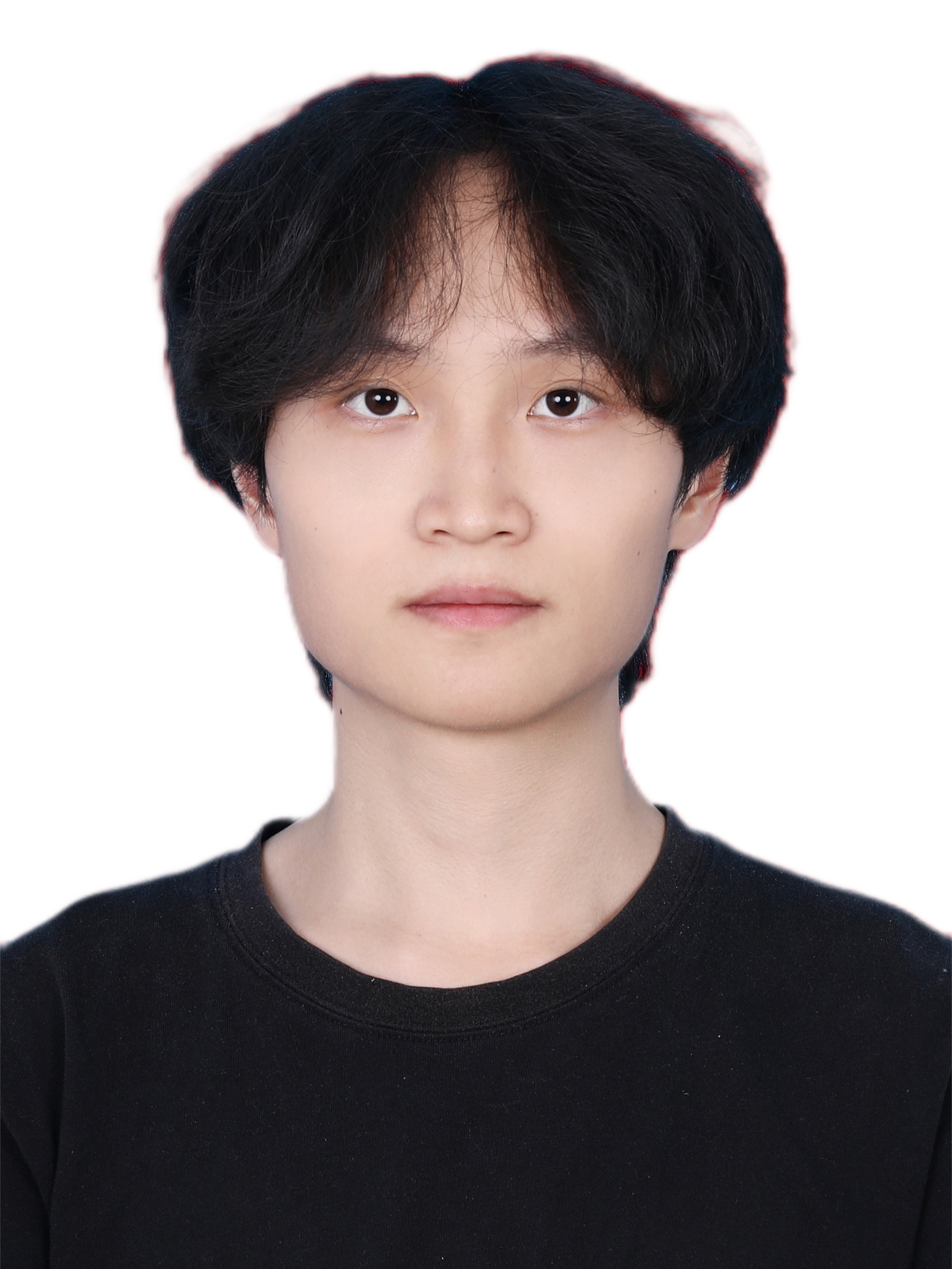}}]{Yongyao Wang}
is a master student at the School of Computer Science and Engineering, Beihang University. His research interests spatiotemporal data mining.
\end{IEEEbiography}
 \vspace{-1.3cm}

\begin{IEEEbiography}[{\includegraphics[width=1in, height=1.25in,clip,keepaspectratio]{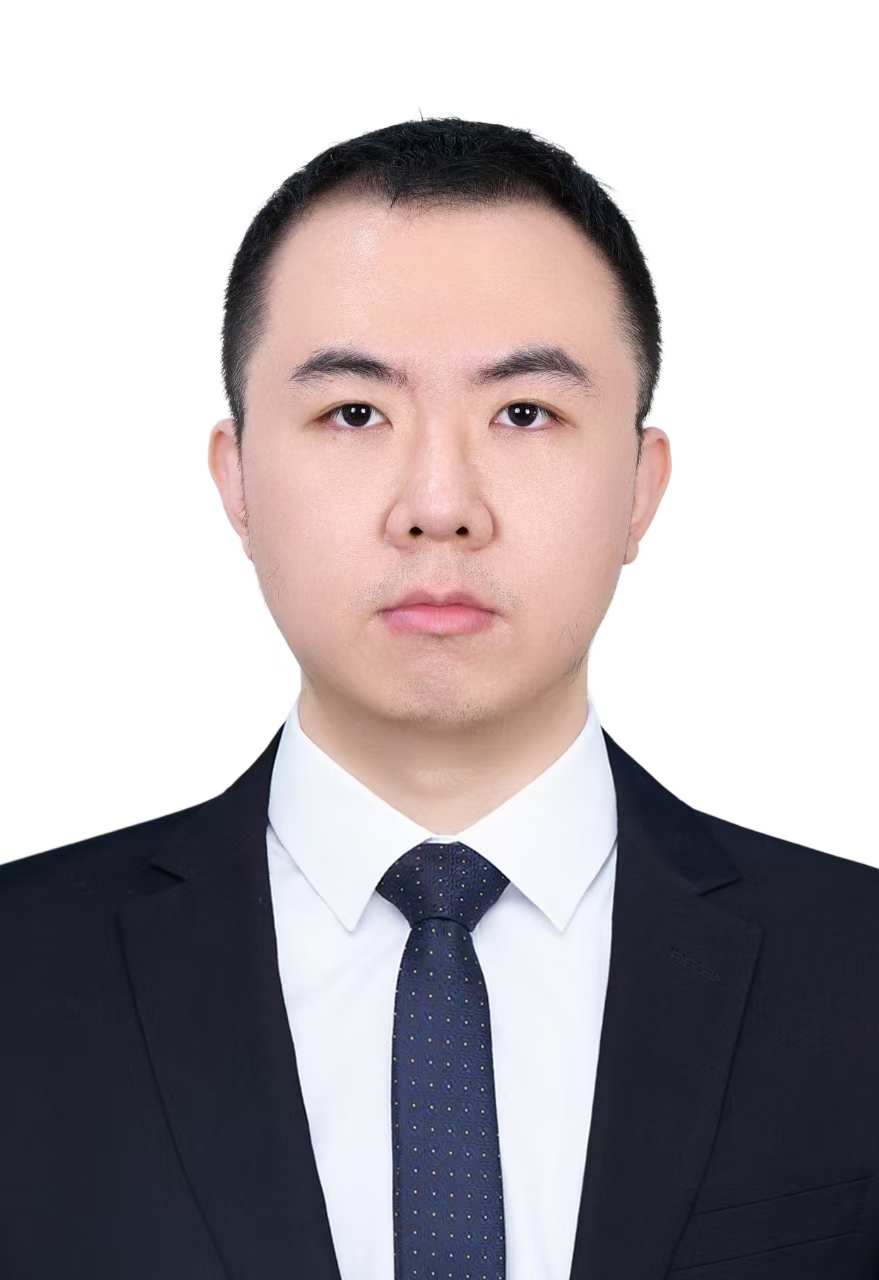}}]{Chen Yang}
is a Ph.D. candidate at the School of Computer Science and Engineering, Beihang University. He received his master's degree from Peking University in 2020. He received his bachelor's degree from Sun Yat-sen University in 2017. His research interests include time series analysis and machine learning in economics.
\end{IEEEbiography}
\vspace{-1.3cm}

\begin{IEEEbiography}[{\includegraphics[width=0.95in, height=1.25in,clip,keepaspectratio]{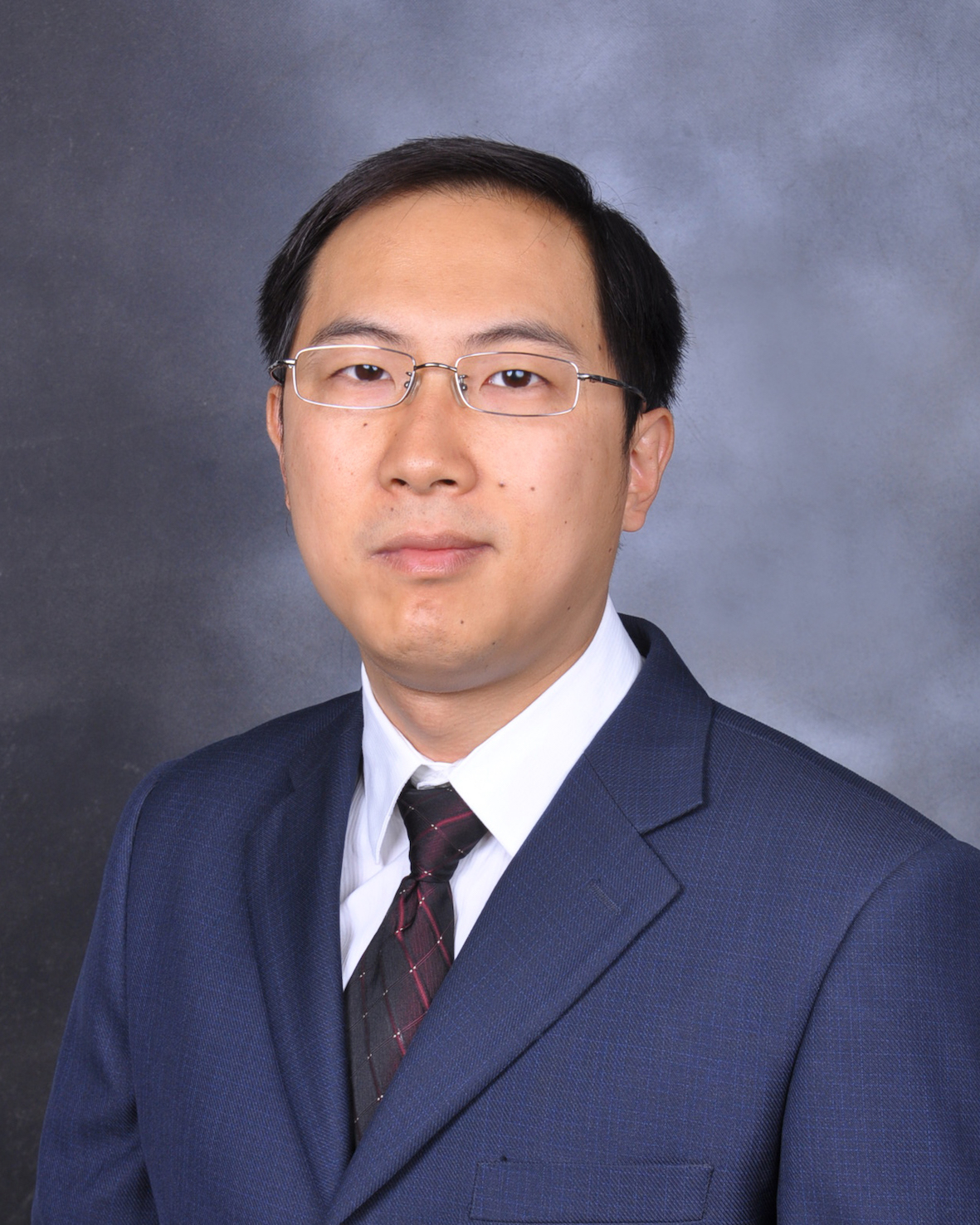}}]{Junjie Wu}
received his Ph.D. degree in Management Science and Engineering from Tsinghua University, from where he also holds a B.E. degree in Civil Engineering. He is currently a full professor and the Dean of the School of Economics and Management, Beihang University. His general area of research is machine learning, with applications in social, urban and financial computing. He is the recipient of the grant of NSFC Distinguished Young Scholars and the grant of Outstanding Young Scientist Program of Beijing Universities.
\end{IEEEbiography}


\end{document}